\documentclass[10pt,journal,compsoc]{IEEEtran}

\usepackage{cite}

\usepackage{graphicx}
\usepackage{amsmath}
\usepackage{acronym}
\usepackage{amssymb}
\usepackage{algorithmic}

\usepackage{subfigure}
\usepackage{subfloat}
\usepackage{multirow}

\usepackage{makecell}
\usepackage{array}
\usepackage{url}
\usepackage{commath}
\usepackage{textcomp}

\usepackage[pagebackref=true,breaklinks=true,colorlinks,bookmarks=false]{hyperref}
\usepackage{marvosym}

\usepackage[symbol]{footmisc}

\usepackage{xspace}
\usepackage{ragged2e}
\usepackage[labelfont=bf, font=footnotesize, justification=justified, format=plain]{caption}

\makeatletter
\DeclareRobustCommand\onedot{\futurelet\@let@token\@onedot}
\def\@onedot{\ifx\@let@token.\else.\null\fi\xspace}

\def\eg{\emph{e.g}\onedot} 
\def\ie{\emph{i.e}\onedot}

\makeatother

\newcommand{\OM}{VRCNet}

\begin{document}
	
	\title{Variational Relational Point Completion Network \\ for Robust 3D Classification}

	\author{Liang Pan$^{1\star}$ \quad
		Xinyi Chen$^{1,2}$ \quad
		Zhongang Cai$^{2,3}$ \quad
		Junzhe Zhang$^{1,2}$ \quad
		\\Haiyu Zhao$^{2,3}$ \quad
		Shuai Yi$^{2,3}$ \quad
		Ziwei Liu$^{1\href{mailto:ziwei.liu@ntu.edu.sg}{\textrm{\Letter}}}$
		\IEEEcompsocitemizethanks{\IEEEcompsocthanksitem Liang Pan, Xinyi Chen, Junzhe Zhang and Ziwei Liu are with the S-Lab, Nanyang Technological University, Singapore,
			639798.
			\IEEEcompsocthanksitem Zhongang Cai, Junzhe Zhang, Haiyu Zhao and Shuai Yi are with SenseTime Research.
			\IEEEcompsocthanksitem Zhongang Cai, Haiyu Zhao and Shuai Yi
			are with Shanghai AI Lab.
			\IEEEcompsocthanksitem The corresponding author is Ziwei Liu: ziwei.liu@ntu.edu.sg
		}
		\thanks{Manuscript received Sep. 9, 2021; revised August 25, 2022.
	}}

	\markboth{IEEE Transactions on Pattern Analysis and Machine Intelligence.}%
	{Shell \MakeLowercase{\textit{et al.}}: Bare Advanced Demo of IEEEtran.cls for IEEE Computer Society Journals}
	
	\IEEEtitleabstractindextext{%
		\begin{abstract}
		\justifying
		Real-scanned point clouds are often incomplete due to viewpoint, occlusion, and noise, which hampers 3D geometric modeling and perception.
		Existing point cloud completion methods tend to generate global shape skeletons and hence lack fine local details.
		Furthermore, they mostly learn a deterministic partial-to-complete mapping, but overlook structural relations in man-made objects.
		To tackle these challenges, this paper proposes a variational framework, \textbf{V}ariational \textbf{R}elational point \textbf{C}ompletion network (\OM{}) with two appealing properties:
		\textbf{1) Probabilistic Modeling.} In particular, we propose a dual-path architecture to enable principled probabilistic modeling across partial and complete clouds.
		One path consumes complete point clouds for reconstruction by learning a point VAE.
		The other path generates complete shapes for partial point clouds, whose embedded distribution is guided by distribution obtained from the reconstruction path during training.
		\textbf{2) Relational Enhancement.} 
		Specifically, we carefully design point self-attention kernel and point selective kernel module to exploit relational point features, which refines local shape details conditioned on the coarse completion.
		In addition, we contribute \textbf{multi-view partial point cloud datasets (MVP and MVP-40 dataset)} containing over 200,000 high-quality scans, which render partial 3D shapes from 26 uniformly distributed camera poses for each 3D CAD model.
		Extensive experiments demonstrate that \OM{} outperforms state-of-the-art methods on all standard point cloud completion benchmarks. 
		Notably, \OM{} shows great generalizability and robustness on real-world point cloud scans.
		Moreover, 
		we can achieve robust 3D classification for partial point clouds with the help of \OM{}, which can highly increase classification accuracy.
		Our project is available at \href{https://paul007pl.github.io/projects/VRCNet}{https://paul007pl.github.io/projects/VRCNet}.
	\end{abstract}
	
	\begin{IEEEkeywords} 
		Point Cloud Completion, 3D Perception, Self-Attention Operations, Multi-View Partial Point Clouds
\end{IEEEkeywords}}

\maketitle

\IEEEdisplaynontitleabstractindextext

\IEEEpeerreviewmaketitle

\ifCLASSOPTIONcompsoc
\IEEEraisesectionheading{\section{Introduction}\label{sec:introduction}}
\else
\section{Introduction}
\label{sec:introduction}
\fi

\IEEEPARstart{P}{oint} cloud is an intuitive representation of 3D scenes and objects, which has extensive applications in various vision and robotics tasks.
Unfortunately, scanned 3D point clouds are usually incomplete owing to occlusions and missing measurements, hampering practical usages, such as 3D modeling and perception.
To mitigate those challenges, it is desirable and important to predict the complete 3D shape from a partially observed point cloud.

The pioneering work PCN~\cite{yuan2018pcn} uses a PointNet-based encoder to generate global features for shape completion, which cannot recover fine geometric details.
The follow-up works~\cite{liu2020morphing,wang2020cascaded,sun2020pointgrow,pan2020ecg,xie2020grnet} provide better completion results by preserving observed geometric details from the incomplete point shape using local features.
However, they~\cite{yuan2018pcn,liu2020morphing,wang2020cascaded,pan2020ecg,xie2020grnet} mostly generate complete shapes by learning a deterministic partial-to-complete mapping, lacking the conditional generative capability based on the partial observation.
Furthermore, 3D shape completion is expected to recover plausible yet fine-grained complete shapes by learning relational structure properties, such as geometrical symmetries, regular arrangements and surface smoothness, which existing methods fail to capture.

\begin{figure}
    \centering
    \includegraphics[width=1\linewidth]{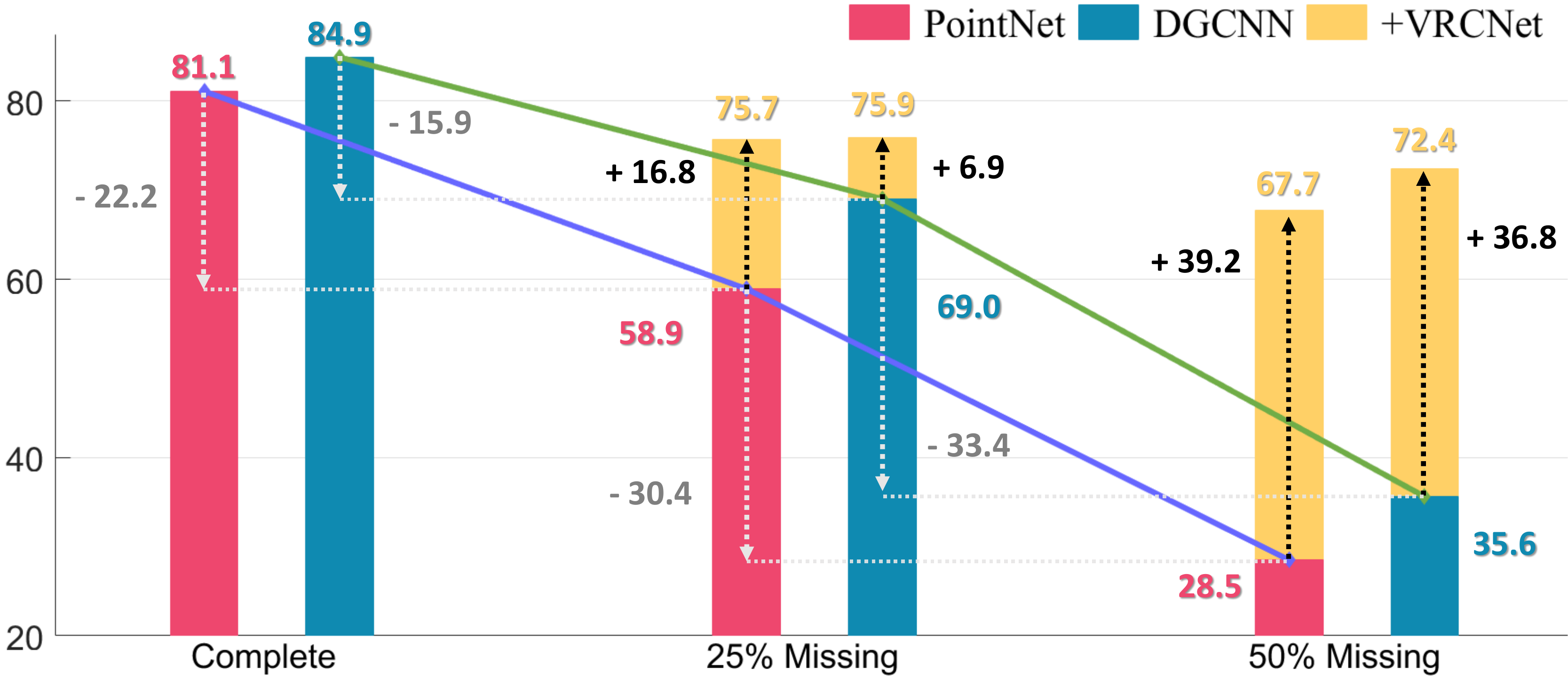}
    \vspace{-7mm}
    \caption{\textbf{
    Point cloud incompleteness impacts classification accuracy on MVP-40 of both PointNet~\cite{qi2017pointnet} ({\color[RGB]{238,71,110}\textbf{red}} bars) and DGCNN~\cite{dgcnn} ({\color[RGB]{15,138,177}\textbf{blue}} bars).}
    The classification accuracy significantly drops when its missing ratio is very large (50\% missing).  After completion by VRCNet, we can highly improve the performance for both methods (see {\color[RGB]{255,208,102}\textbf{yellow}} bars).}
    \label{fig:cp_pc}
\end{figure}
\begin{figure*}
    \centering
    \includegraphics[width=.975\textwidth]{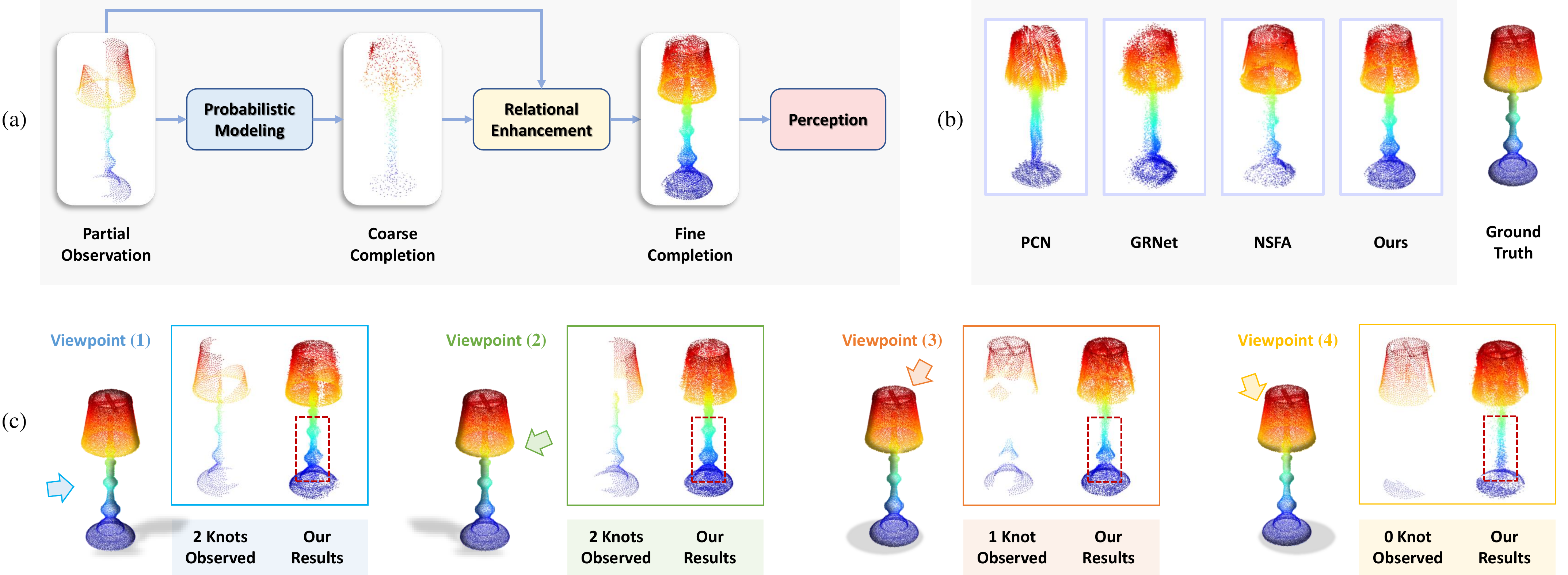}
    \vspace{-1.5mm}
    \caption{(a) \textbf{System Overview.} \OM{} is firstly used for point cloud completion with two consecutive stages: probabilistic modeling and relational enhancement, which facilitates downstream perception tasks.  
    (b) \textbf{Qualitative Results} show that \OM{} generates better shape details than the other works~\cite{yuan2018pcn,xie2020grnet,zhang2020detail}.
    (c) \textbf{Our completion results conditioned on partial observations.} 
    The arrows indicate the viewing angles. 
    In (1) and (2), 2 knots are partially observed for the pole of the lamp, and hence we generate 2 complete knots.
    In (3), only 1 knot is observed, and then we reconstruct 1 complete knot. 
    If no knots are observed (see (4)), \OM{} generates a smooth pole without knots.}
    \label{fig:first_glance}
\end{figure*}

To this end, we propose Variational Relational Point Completion network (entitled as \OM), which consists of two consecutive encoder-decoder sub-networks that serve as ``probabilistic modeling'' (PMNet) and ``relational enhancement'' (RENet), respectively (shown in Fig.~\ref{fig:first_glance}~(a)).
The first sub-network, PMNet, embeds global features and latent distributions from incomplete point clouds, and predicts the overall skeletons (\ie coarse completions, see Fig.~\ref{fig:first_glance} (a)) that are used as 3D adaptive anchor points for exploiting multi-scale point relations in RENet.
Inspired by~\cite{zheng2019pluralistic}, PMNet uses smooth complete shape priors to improve the generated coarse completions using a dual-path architecture consisting of two parallel paths: 1) a reconstruction path for complete point clouds, and 2) a completion path for incomplete point clouds.
During training, we regularize the consistency between the encoded posterior distributions from partial point clouds and the prior distributions from complete point clouds.
With the help of the generated coarse completions, the second sub-network RENet strives to enhance structural relations by learning multi-scale local point features.
Motivated by the success of local relation operations in image recognition~\cite{zhao2020exploring,hu2019local}, we propose the Point Self-Attention Kernel (PSA) as a basic building block for RENet.
Instead of using fixed weights, PSA interleaves local point features by adaptively predicting weights based on the learned relations among neighboring points.
Inspired by the Selective Kernel (SK) unit~\cite{li2019selective}, we propose the Point Selective Kernel Module (PSK) that utilizes multiple branches with different kernel sizes to exploit and fuse multi-scale point features, which further improves the performance.

Moreover, we create large-scale Multi-View Partial point cloud  datasets (MVP and MVP-40) with over 200,000 high-quality scanned partial and complete point clouds.
For each complete 3D CAD model selected from ShapeNet~\cite{shapenet2015} and ModelNet~\cite{wu20153d}, we randomly render 26 partial point clouds from uniformly distributed camera views on a unit sphere, which improves the data diversity.
Experimental results on our MVP and Completion3D benchmark~\cite{tchapmi2019topnet} show that \OM{} outperforms SoTA methods.
In Fig.~\ref{fig:first_glance} (b), \OM{} reconstructs richer details than the other methods by implicitly learning the shape symmetry from this incomplete lamp.
Given different partial observations, \OM{} can predict different plausible complete shapes (Fig.~\ref{fig:first_glance} (c)).
Furthermore, \OM{} can generate impressive complete shapes for incomplete real-world scans from KITTI~\cite{Geiger2012CVPR} and ScanNet~\cite{dai2017scannet}, which reveals its remarkable robustness and generalizability.

In view that 3D perception degrades with point cloud incompleteness,
we study two representative 3D perception networks, PointNet~\cite{qi2017pointnet} and DGCNN~\cite{dgcnn}, for classifying incomplete point clouds and their completion results.
As shown in Fig.~\ref{fig:cp_pc}, classification performance of both DGCNN and PointNet can be influenced by point cloud incompleteness, especially for those with large missing ratios.
With the help of completion by \OM{}, we can highly improve their classification accuracy.
Comparing against the other completion networks, \OM{} achieves much better overall perception improvements, which further validates that \OM{} generates better complete 3D point clouds, and then conducts robust 3D perception against incompleteness. 

The key contributions can be summarized as:
\begin{itemize}
    \item We propose a novel \textbf{V}ariational \textbf{R}elational point \textbf{C}ompletion \textbf{Net}work (\OM), which  first performs probabilistic modeling using a novel dual-path network followed by a relational enhancement network.
    \item We design multiple relational modules that can effectively exploit and fuse multi-scale point features for point cloud analysis, such as the Point Self-Attention Kernel and the Point Selective Kernel Module.
    \item We study different 3D perception networks on classifying incomplete point clouds with random missing parts, and completion by \OM{} shows significant perception improvements.
    \item Furthermore, we contribute large-scale multi-view partial point cloud datasets (MVP and MVP-40) with over 200,000 high-quality 3D point shapes. 
\end{itemize}
Extensive experiments show that \OM{} outperforms previous SoTA methods on all evaluated benchmark datasets.

\begin{figure*}
    \centering
    \includegraphics[width=.975\linewidth]{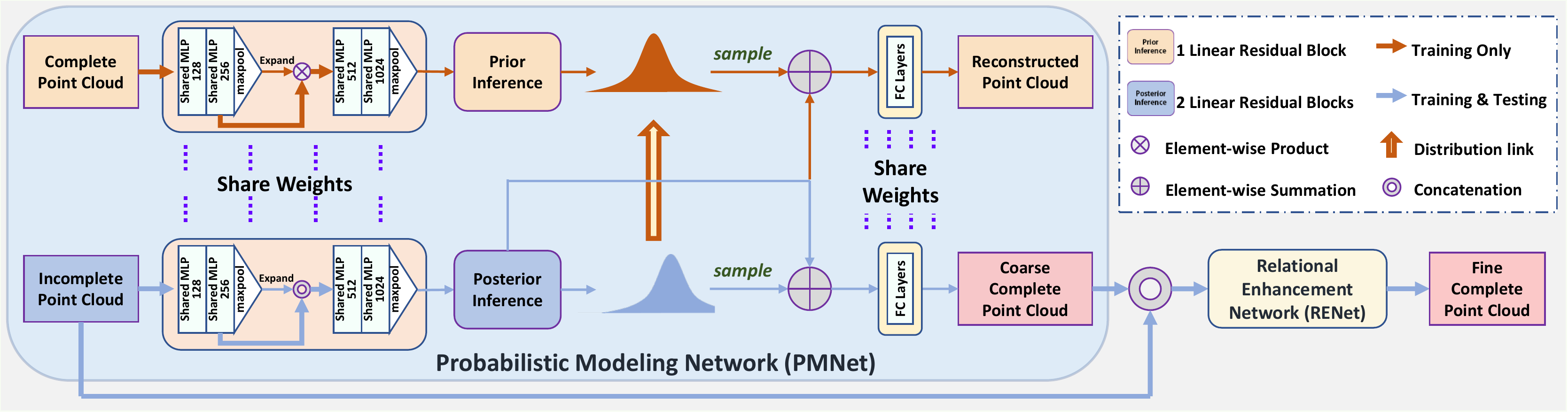}
    \vspace{-2mm}
    \caption{
    \textbf{
    PMNet (light blue block) consists of two parallel paths}, the upper construction path (orange line) and the lower completion path (blue line). The reconstruction path is only used in training, and the completion path generates a coarse completion based on the inferred distribution and global features. Subsequently, RENet (Fig \ref{fig:spsakn}) adaptively exploits relational structure properties to predict the fine complete point cloud.}
    \label{fig:overview}
\end{figure*}

\section{Related Work}
In this section, we summarize recent research advances with respect to the following three aspects:
1) deep learning on 3D point cloud, especially multi-scale point features extraction;
2) complete point cloud generation based on partially observed point clouds;
3) object category recognition for incomplete point clouds.

\noindent\textbf{Multi-scale Features Exploitation.}
Convolutional operations have yielded impressive results for image applications~\cite{krizhevsky2017imagenet,he2016deep,simonyan2014very}.
However, conventional convolutions cannot be directly applied to point clouds due to the absence of regular grids.
Previous networks mostly exploit local point features by two operations: local pooling~\cite{wang2019dynamic,qi2017pointnet++,pan2019pointatrousgraph} and flexible convolution~\cite{groh2018flex,thomas2019kpconv,li2018pointcnn,wu2019pointconv}.
Self-attention often uses linear layers, such as fully-connected (FC) layers and shared multilayer perceptron (shared MLP) layers, which are appropriate for point clouds.
In particular, recent works~\cite{zhao2020exploring,hu2019local,parmar2019stand} have shown that local self-attention (\ie relation operations) can outperform their convolutional counterparts, which holds the exciting prospect of designing networks for point clouds.
Similar to self-attention operations, Zhao et. al.~\cite{zhao2020point} and Guo et. al.~\cite{guo2020pct} use transformer-based operations to learn point feature relations, and they provide impressive performance on various high-level perception-based applications, such as classification and segmentation.
Therefore, it is interesting to study relational operations for the low-level 3D point shape completion task.

\noindent\textbf{Point Cloud Completion.}
Point cloud completion targets at recovering a complete 3D shape based on a partial point cloud observation.
Earlier works~\cite{li2016shape,wu20153d,dai2017shape} often leverage volumetric representations for 3D shape completion. 
However, large memory requirements of voxel grids highly limits its capability in reconstructing high-quality local structures.
Recently, many research works directly consume partial point clouds for completion.
PCN~\cite{yuan2018pcn} 
first generates a coarse completion based on learned global features from the partial input point cloud, which is upsampled using folding operations~\cite{yang2018foldingnet}.
Following PCN, TopNet~\cite{tchapmi2019topnet} proposes a tree-structured decoder to predict complete shapes.
To preserve and recover local details, previous approaches~\cite{wang2020cascaded,pan2020ecg,xie2020grnet} exploit local features to refine their 3D completion results.
NSFA~\cite{zhang2020detail} recovers complete 3D shapes by combining known features and missing features.
However, NSFA assumes that the ratio of the known part and the missing part is around $1:1$ (\ie, the visible part should be roughly a half of the whole object), which does not hold for point clouds completion in most cases.
{Most recently, PoinTr~\cite{yu2021pointr} shows impressive completion results by adopting a transformer encoder-decoder architecture, and it also reveals the effectiveness of attention operations for point cloud completion.
CP3~\cite{xu2022cp3} proposes a generic pretrain-prompt-predict paradigm to enhance the semantic awareness for point cloud completion.
}

\noindent\textbf{Partial Point Cloud Perception.}
According to Marr's theory~\cite{marr1979computational}, human vision is accustomed to recognizing view-centered partial shapes (2.5D) to observe the 3D world.
Therefore, many research works~\cite{wu20153d, su2015render, su2015multi, Qi_2016_CVPR, jaritz2019multi} are proposed to recognize 3D objects by rendering different incomplete observations from multiple viewpoints.
For example, 3D ShapeNet~\cite{wu20153d} studies generic shape representation for jointly hallucinating missing structures and predicting object categories.
Following the success of PointNet~\cite{qi2017pointnet}, many recent research works~\cite{dgcnn, pan2019pointatrousnet, guo2020pct, zhao2020point} directly perform deep learning on 3D points.
However, they mostly perform perception on clean and complete 3D shapes while overlooking the incompleteness nature of real-scans, which can highly limit their performance for real applications.
{Recently, Ren et. al.~\cite{ren2022benchmarking} benchmark and analyze point cloud classifier robustness under various corruptions, such as incompleteness (drop) and noise (jitter).}
In this work, we resolve the perception challenge for incomplete point clouds by first conducting point cloud completion before recognition.
\begin{figure*}
    \centering
    \includegraphics[width=0.95\linewidth]{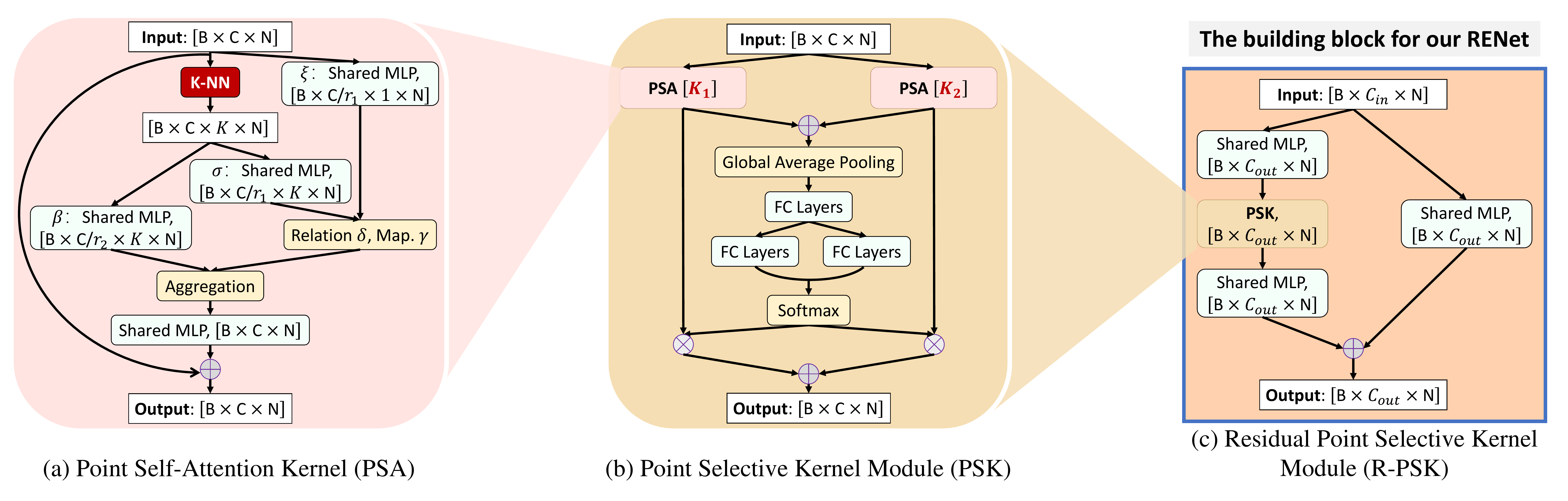}
    \vspace{-4mm}
    \caption{\textbf{Our proposed point kernels.} (a) Our PSA adaptively aggregate neighboring point features. (b) Using selective kernel unit, our PSK can adaptively adjust receptive fields to exploit and fuse multi-scale point features. (c) By adding a residual connection, we construct our RPSK that is an important building block for our RENet.}
    \vspace{-2mm}
    \label{fig:point_modules}
\end{figure*}
\section{Our Approach}
\subsection{Problem Formulation}
Real-scanned point clouds are mostly incomplete, which challenges low-level 3D modeling and high-level 3D perception.
To tackle these challenges, we firstly focus on predicting fine-grained complete point clouds for partial observations.
Afterwards, we study favorable improvements by point cloud completion for classification,
which leads to robust 3D perception for incomplete point clouds.

We define the incomplete point cloud $\mathbf{X}$ as a partial observation for a 3D object, and a complete point cloud $\mathbf{Y}$ is sampled from the surfaces of the object. 
Note that $\mathbf{X}$ need not to be a subset of $\mathbf{Y}$, since $\mathbf{X}$ and $\mathbf{Y}$ are generated by two separate processes.
The point cloud completion task aims to predict a complete shape $\mathbf{Y}^{\prime}$ conditioned on $\mathbf{X}$.
\OM{} generate high-quality complete point clouds in a coarse-to-fine fashion.
Firstly, we predict a coarse completion $\mathbf{Y}^{\prime}_{c}$ based on embedded global features and an estimated parametric distribution.
Subsequently, we recover relational geometries for the fine completion $\mathbf{Y}^{\prime}_{f}$ by exploiting multi-scale point features with novel self-attention modules.

\subsection{VRCNet for Point Cloud Completion}
\subsubsection{Probabilistic Modeling}
Previous networks~\cite{yuan2018pcn,tchapmi2019topnet} tend to decode learned global features to predict overall shape skeletons as their completion results, which cannot recover fine-grained geometric details.
However, it is still beneficial to first predict the shape skeletons before refining local details for the following reasons: 1) shape skeletons describe the coarse complete structures, especially for those areas that are entirely missing in the partial observations;
2) shape skeletons can be regarded as adaptive 3D anchor points for exploiting local point features in incomplete point clouds~\cite{pan2020ecg}.
With these benefits, we propose the Probabilistic Modeling network (PMNet) to generate the overall skeletons (\ie coarse completions) for incomplete point clouds.

In contrast to previous methods, PMNet employs probabilistic modeling to predict the coarse completions based on both embedded global features and learned latent distributions.
Moreover, we employ a dual-path architecture (shown in Fig.~\ref{fig:overview}) that contains two parallel pipelines: the upper reconstruction path for complete point clouds $\mathbf{Y}$ and the lower completion path for partial point clouds $\mathbf{X}$.
The reconstruction path follows a variational auto-encoder (VAE) scheme. 
It first encodes global features $\mathbf{z_g}$ and latent distributions $\,q_{\phi}(\mathbf{z_g}|\mathbf{Y})$ for the complete shape $\mathbf{Y}$, and then it uses a decoding distribution $p_{\theta}^r(\mathbf{Y}|\mathbf{z_g})$ to recover a complete shape $\mathbf{Y}_r^{\prime}$.
The objective function for the reconstruction path can be formulated as:
\begin{equation}
    \begin{aligned}
        \mathcal{L}_{rec} = &-\lambda \, \mathbf{KL}\big[q_{\phi}(\mathbf{z_g}|\mathbf{Y}) \, \big\| \,p(\mathbf{z_g})\big] \\
        &+ \mathbb{E}_{p_{data}(\mathbf{Y})}\mathbb{E}_{q_{\phi}(\mathbf{z_g}|\mathbf{Y})}\big[\log {p_{\theta}^r}(\mathbf{Y}|\mathbf{z_g})\big],
    \end{aligned}
\end{equation}
where $\mathbf{KL}$ is the KL divergence, $\mathbb{E}$ denotes the estimated expectations of certain functions, $p_{data}(\mathbf{Y})$ denotes the true underlying distribution of data, and $p(\mathbf{z_g})=\mathcal{N}(\mathbf{0}, \mathbf{I})$ is the conditional prior predefined as a Gaussian distribution, and $\lambda$ is a weighting parameter.

The completion path has a similar structure as the constructive path, and both two paths share weights for their encoder and decoder except the distribution inference layers.
Likewise, the completion path aims to reconstruct a complete shape $\mathbf{Y}^{\prime}_c$ based on global features $\mathbf{z_g}$ and latent distributions $p_{\psi}(\mathbf{z_g}|\mathbf{X})$ from an incomplete input $\mathbf{X}$.
To exploit the most salient features from the incomplete point cloud, we use the learned conditional distribution $q_{\phi}(\mathbf{z_g}|\mathbf{Y})$ encoded by its corresponding complete 3D shapes $\mathbf{Y}$ to 
regularize
latent distributions $p_{\psi}(\mathbf{z_g}|\mathbf{X})$ during training (shown as the Distribution Link in Fig.~\ref{fig:overview}, the arrow indicates that we regularize $p_{\psi}(\mathbf{z_g}|\mathbf{X})$ to approach $q_{\phi}(\mathbf{z_g}|\mathbf{Y})$).
Hence, $q_{\phi}(\mathbf{z_g}|\mathbf{Y})$ constitutes the prior distributions, $p_{\psi}(\mathbf{z_g}|\mathbf{X})$ is the posterior importance sampling function, and the objective function for completion path is defined as follows:
\begin{equation}
    \begin{aligned}
        \mathcal{L}_{com} = &-\lambda \, \mathbf{KL}\big[q_{\phi}(\mathbf{z_g}|\mathbf{Y}) \, \big\| \,p_{\psi}(\mathbf{z_g}|\mathbf{X})\big] \\
        &+ 
        \mathbb{E}_{p_{data}(\mathbf{X})}\mathbb{E}_{p_{\psi}(\mathbf{z_g}|\mathbf{X})}\big[\log {p_{\theta}^c}(\mathbf{Y}|\mathbf{z_g})\big],
    \end{aligned}
\end{equation}
where $\phi$, $\psi$ and $\theta$ represent different network weights of their corresponding functions.
Notably, the reconstruction path is only used in training, and hence the dual-path architecture does not influence our inference efficiency.

\subsubsection{Relational Enhancement}
After obtaining coarse completions $\mathbf{Y}^{\prime}_c$, the Relational Enhancement network (RENet) targets at enhancing structural relations to recover local shape details.
Although previous methods~\cite{wang2020cascaded,zhang2020detail,pan2020ecg} can preserve observed geometric details by exploiting local point features, they cannot effectively extract structural relations (\eg geometric symmetries) to recover those missing parts conditioned on the partial observations.
Inspired by the relation operations for image recognition~\cite{hu2019local,zhao2020exploring}, we propose the Point Self-Attention kernel (PSA) to adaptively aggregate local neighboring point features with learned relations in neighboring points (Fig.~\ref{fig:point_modules} (a)). 
The operation of PSA is formulated as:
\begin{equation}
    \mathbf{y}_i = \underset{j\in \mathcal{N}(i)}{\sum}\alpha (\mathbf{x}_{\mathcal{N}(i)})_j \, \odot \beta(\mathbf{x}_j),
\end{equation}
where $\mathbf{x}_{\mathcal{N}(i)}$ is the group of point feature vectors for the selected K-Nearest Neighboring (K-NN) points $\mathcal{N}(i)$.
$\alpha (\mathbf{x}_{\mathcal{N}(i)})$ is a weighting tensor for all selected feature vectors.
$\beta(\mathbf{x}_j)$ is the transformed features for point $j$, which has the same spatial dimensionality with $\alpha (\mathbf{x}_{\mathcal{N}(i)})_j$.
Afterwards, we obtain the output $\mathbf{y}_i$ using an element-wise product $\odot$, which performs a weighted summation for all points $j\in \mathcal{N}(i)$.
The weight computation $\alpha (\mathbf{x}_{\mathcal{N}(i)})$ can be decomposed as follows:
\begin{equation}
    \begin{aligned}
        \alpha (\mathbf{x}_{\mathcal{N}(i)}) &= \gamma\big(\delta(\mathbf{x}_{\mathcal{N}(i)})\big), \\
        \delta(\mathbf{x}_{\mathcal{N}(i)}) &= 
        \big[\sigma(\mathbf{x}_i), [\xi(\mathbf{x}_j)]_{\forall j \in \mathcal{N}(i)}\big],
    \end{aligned}
\end{equation}
where $\gamma$, $\sigma$ and $\xi$ are all shared MLP layers (Fig.~\ref{fig:point_modules} (a)), and the relation function $\delta$ 
combines all feature vectors $\mathbf{x}_j \in \mathbf{x}_{\mathcal{N}(i)}$ by using concatenation operations.

Observing that different relational structures can have different scales, we enable the neurons to adaptively adjust their receptive field sizes by leveraging the selective kernel unit~\cite{li2019selective}. 
Hence, we construct the Point Selective Kernel module (PSK), which adaptively fuses learned structural relations from different scales.
In Fig.~\ref{fig:point_modules} (b), we show a two-branch case, which has two PSA kernels with different kernel (\ie K-NN) sizes.
The operations of the PSK are formulated as:
\begin{equation}
\hspace{-2mm}
\left\{
    \begin{aligned}
        & \mathbf{V}_c = \mathbf{\tilde{U}}_c \cdot a_c + \mathbf{\hat{U}}_c \cdot b_c\, , \\
        & a_c = \frac{e^{\mathbf{A}_c\mathbf{z}}}{e^{\mathbf{A}_c\mathbf{z}} + e^{\mathbf{B}_c\mathbf{z}}}, \quad b_c = \frac{e^{\mathbf{B}_c\mathbf{z}}}{e^{\mathbf{A}_c\mathbf{z}} + e^{\mathbf{B}_c\mathbf{z}}},  \\
        & \mathbf{U} = \mathbf{\tilde{U}} +  \mathbf{\hat{U}}, \quad s_c = 
        \frac{1}{{N}}\sum_{i=1}^{{N}}\mathbf{U}_c(i), \quad
        \mathbf{z} = \eta(\mathbf{W}\mathbf{s}),
    \end{aligned}
\right .
\end{equation}
where $\mathbf{\hat{U}}, \mathbf{\tilde{U}} \in \mathbb{R}^{N\times C}$ are point features encoded by two kernels respectively, $\mathbf{\tilde{V}} \in \mathbb{R}^{N\times C}$ is the final fused features, $\mathbf{s}$ is obtained by using element-wise average pooling over all $N$ points for each feature $c\in C$, $\eta$ is a FC layer, $\mathbf{W}\in \mathbb{R}^{d\times C}$, $\mathbf{A},\mathbf{B}\in \mathbb{R}^{C\times d}$, and $d$ is a reduced feature size.

\begin{figure}
    \centering
    \includegraphics[width=1\linewidth]{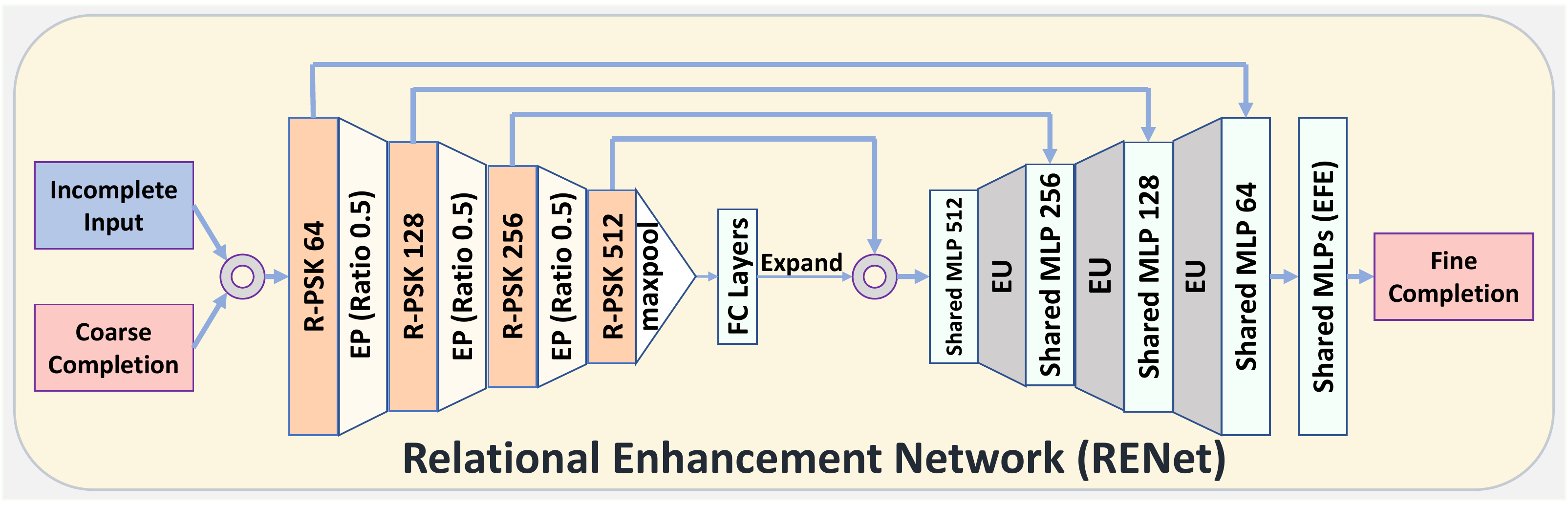}
    \vspace{-5mm}
    \caption{\textbf{Our Relational Enhancement Network (RENet)} uses a hierarchical encoder-decoder architecture, which effectively learns multi-scale structural relations.}
    \label{fig:spsakn}
\end{figure}

Furthermore, we add an residual path besides the main path (shown in Fig.~\ref{fig:point_modules} (c)) and then construct the Residual Point Selective Kernel module (R-PSK) that is used as a building block for RENet.
As shown in Fig.~\ref{fig:spsakn}, RENet follows a hierarchical encoder-decoder architecture by using Edge-preserved Pooling (EP) and Edge-preserved Unpooling (EU) modules~\cite{pan2019pointatrousgraph}. 
We use an Edge-aware Feature Expansion (EFE) module~\cite{pan2020ecg} to expand point features, which generates high-resolution complete point clouds with predicted fine local details.
Consequently, multi-scale structural relations can be exploited for fine details generation.

\subsubsection{Loss Functions}
Our \OM{} is trained end-to-end, and the training loss consists of three parts: $\mathcal{L}_{rec}$ (reconstruction path), $\mathcal{L}_{com}$ (completion path) and $\mathcal{L}_{fine}$ (relational enhancement).
$\mathcal{L}_{rec}$ and $\mathcal{L}_{com}$ have two loss items, a $\mathbf{KL}$ divergence loss and a reconstruction loss, while $\mathcal{L}_{fine}$ only has a reconstruction loss.
The KL divergence is defined as:
\begin{equation}
    \mathcal{L}_{\mathbf{KL}}(q, p) = -\mathbf{KL} \big[q(\mathbf{z})\,\big\|\,p(\mathbf{z})\big].
\end{equation}
Considering the training efficiency, we choose the symmetric Chamfer Distance (CD) as the reconstruction loss:
\begin{equation}
    \mathcal{L}_{\mathbf{CD}}(\mathbf{P}, \mathbf{Q}) = \frac{1}{|\mathbf{P}|}\sum_{x\in\mathbf{P}}\underset{y\in\mathbf{Q}}{\min}\|x-y\|^2 + \frac{1}{|\mathbf{Q}|}\sum_{y\in\mathbf{Q}}\underset{x\in\mathbf{P}}{\min}\|x-y\|^2,
    \label{eq:cd_loss}
\end{equation}
where $x$ and $y$ denote points that belong to two point clouds $\mathbf{P}$ and $\mathbf{Q}$, respectively.
Consequently, the joint loss function can be formulated as:
\begin{equation}
\begin{aligned}
    \mathcal{L} = & \lambda_{rec}\mathcal{L}_{rec} + \lambda_{com}\mathcal{L}_{com} + \lambda_{fine}\mathcal{L}_{fine} \\
    = &\lambda_{rec}\big[ \mathcal{L}_\mathbf{KL}(q_{\phi}(\mathbf{z_g}|\mathbf{Y}),\, \mathcal{N}(\mathbf{0}, \mathbf{I})) + \mathcal{L}_\mathbf{CD}(\mathbf{Y}_r^{\prime}, \mathbf{Y}) \big] \\
    + &\lambda_{com}\big[ \mathcal{L}_\mathbf{KL}(p_{\psi}(\mathbf{z_g}|\mathbf{X}), \, q_{\phi}(\mathbf{z_g}|\mathbf{Y})) + \mathcal{L}_{\mathbf{CD}}(\mathbf{Y}_c^{\prime}, \mathbf{Y}) \big] \\
    + &\lambda_{fine}\mathcal{L}_\mathbf{CD}(\mathbf{Y}_f^{\prime}, \mathbf{Y}),
\end{aligned}
\end{equation}
where $\lambda_f$, $\lambda_r$ and $\lambda_c$ are the weighting parameters.

\subsection{Point Cloud Completion for Classification}
Point cloud classification predicts the object category by encoding representations for describing 3D geometry.
Recent 3D classification networks~\cite{qi2017pointnet,dgcnn} mostly deal with complete point clouds, but their accuracy can be largely impacted while partial point cloud incompleteness increasing (see Fig.~\ref{fig:cp_pc}).
Point cloud completion and classification for incomplete point clouds can jointly benefit each other~\cite{wu20153d}, and hence high-quality complete point clouds should give rise to better classification performance.
However, few research works are proposed to study both classification and completion for unorganized partial point clouds.
In light of this, we focus on robust 3D classification for partial point clouds, which can benefit from completion methods.

\begin{figure}
    \centering
    \includegraphics[width=1\linewidth]{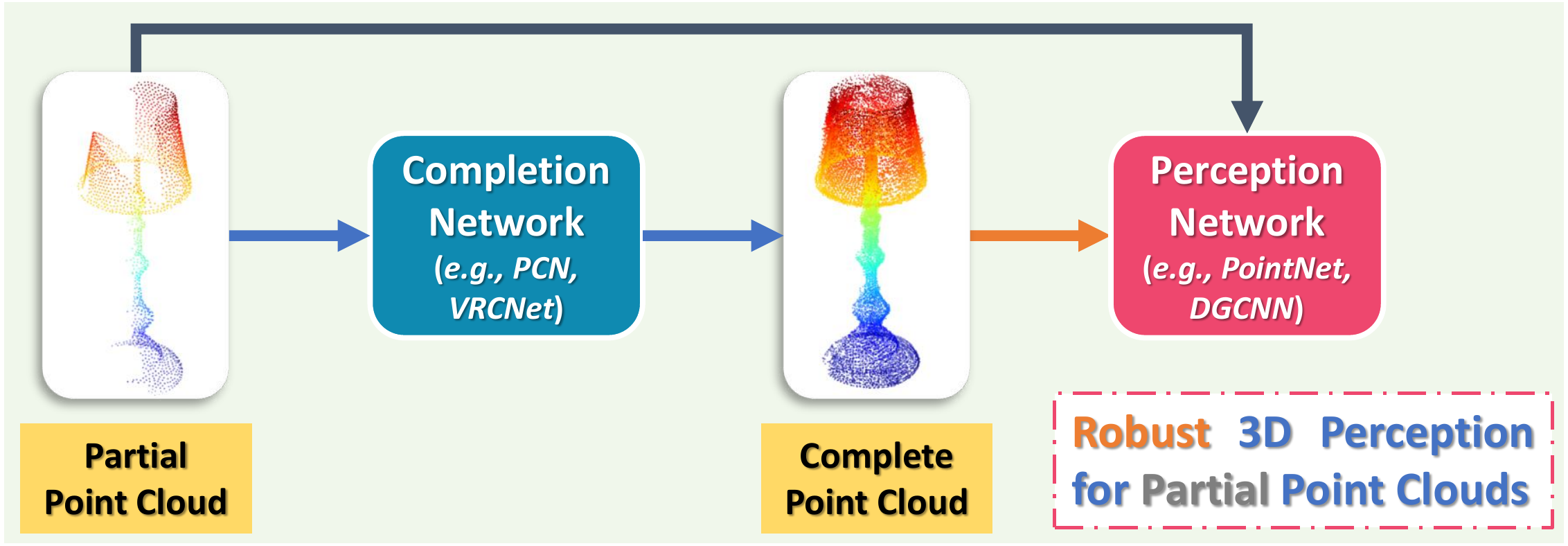}
    \vspace{-5mm}
    \caption{\textbf{Point cloud Completion for Perception.}  Various point cloud completion networks are compared with different perception networks.}
    \label{fig:pc_block}
\end{figure}

Two representative classification networks, PointNet~\cite{qi2017pointnet} and DGCNN~\cite{dgcnn}, are employed for classifying different completion results. 
PointNet focuses on learning global point feature embeddings, while DGCNN uses neighboring point graphs for exploiting local geometric features.
To resolve the challenge that their classification performance can drop caused by shape incompleteness, we first conduct the complete point cloud prediction conditioned on the observed partial point clouds.
Note that no category information is provided during completion, and hence all evaluated completion methods use low-level structural information only.
As shown in Fig.~\ref{fig:pc_block}, we investigate the classification performance improvements with the predicted complete point clouds by different completion methods in comparison against original partial point cloud inputs.
On the other hand, classification accuracy can further validate completion quality by different methods, in terms of global point feature representations (PointNet) and local point distributions (DGCNN), respectively.
Considering both global skeletons and local details, the better quality of the generated point cloud completion results, the higher classification accuracy mostly should be achieved.  
Comprehensive experiments and analysis are reported in Sec.~\ref{sec:Exp}.

\setlength{\tabcolsep}{4.7pt}
\begin{table*}[]
    \caption{
    \textbf{Comparing MVP with existing datasets.} 
    MVP has many appealing properties:
    1) diversity of uniform views; 2) large-scale and high-quality; 3) rich categories. 
    Note that both PCN and C3D only randomly render \textbf{One} incomplete point cloud for each CAD model to construct their testing sets.
    (C3D: Completion3D; Cat.: Categories; Distri.: Distribution; Reso.: Virtual Camera Resolution or Missing Ratios of Incomplete Point Clouds; PC: Point Cloud; FPS: Farthest Point Sampling; PDS: Poisson Disk Sampling. Point cloud resolution is shown as multiples of 2048 points.)
    }
    \vspace{-4mm}
    \begin{center}
    \small{
        \begin{tabular}{l|c|cc|cc|ccc|cc|cc}
            \Xhline{1pt}
             \multirow{2}{*}{} &
             \multirow{2}{*}{\#Cat.} & 
             \multicolumn{2}{c|}{Training Set} &
             \multicolumn{2}{c}{Testing Set} &
             \multicolumn{3}{|c|}{Virtual Camera} & 
             \multicolumn{2}{c|}{Complete PC} &
             \multicolumn{2}{c}{Incomplete PC} \\
             & & \#CAD & \#Pair & \#CAD & \#Pair & Num. & Distri. & Reso. & 
             Sampling & Reso. & Sampling & Reso. \\
            \hline\hline 
             \small PCN~\cite{yuan2018pcn} & \scriptsize 8 & \scriptsize 28974 & \scriptsize $\sim$200k & \scriptsize 1200 & \scriptsize 1200 & \scriptsize 8 & \scriptsize Random & \scriptsize {160\texttimes120} & \scriptsize Uniform & \scriptsize 8\texttimes & \scriptsize Random & \scriptsize $\sim$3000 \\ 
             \small C3D~\cite{tchapmi2019topnet} & \scriptsize 8 & \scriptsize 28974 & \scriptsize 28974 & \scriptsize 1184 & \scriptsize 1184 & \scriptsize 1 & \scriptsize Random & \scriptsize {160\texttimes120} & \scriptsize Uniform & \scriptsize 1\texttimes & \scriptsize Random & \scriptsize 1\texttimes \\ 
            \small MSN~\cite{liu2020morphing} & \scriptsize 8 & \scriptsize 28974 & \scriptsize $\sim$1.4m & \scriptsize 1200 & \scriptsize 1200 & \scriptsize 50 & \scriptsize Random & \scriptsize {160\texttimes120} & \scriptsize Uniform & \scriptsize 4\texttimes & \scriptsize Random & \scriptsize $\sim$5000 \\
            \small Wang et. al.~\cite{wang2020cascaded} & \scriptsize 8 & \scriptsize 28974 & \scriptsize 28974 & \scriptsize 1200 & \scriptsize 1200 & \scriptsize 1 & \scriptsize Random & \scriptsize {160\texttimes120} & \scriptsize Uniform & \scriptsize 1\texttimes & \scriptsize Random & \scriptsize 1\texttimes \\
            \small SANet~\cite{wen2020point} & \scriptsize 8 & \scriptsize 28974 & \scriptsize $\sim$200k & \scriptsize 1200 & \scriptsize 1200 & \scriptsize 8 & \scriptsize Random & \scriptsize {160\texttimes120} & \scriptsize Uniform & \scriptsize 1\texttimes & \scriptsize Random & \scriptsize 1\texttimes \\
            \small NSFA~\cite{zhang2020detail} & \scriptsize 8 & \scriptsize 28974 & \scriptsize $\sim$200k & \scriptsize 1200 & \scriptsize 1200 & \scriptsize 7 & \scriptsize Random & \scriptsize {160\texttimes120} & \scriptsize Uniform & \scriptsize 8\texttimes & \scriptsize Random & \scriptsize 1\texttimes \\
            \hline
             MVP & \footnotesize \textbf{16} & \footnotesize 2400 & \footnotesize  62400  & \footnotesize 1600 & {41600} & \footnotesize {26} & \footnotesize \textbf{Uniform} & \footnotesize \textbf{{1600\texttimes1200}} & \footnotesize \textbf{PDS} & \footnotesize \textbf{1/2/4/8\texttimes} & 
             \footnotesize \textbf{FPS} & \footnotesize 1\texttimes \\
             MVP-40 & \footnotesize \textbf{40} & \footnotesize 1600 & \footnotesize  41600  & \footnotesize 2468 & \textbf{64168} & \footnotesize {26} & \footnotesize \textbf{Uniform} & \footnotesize{25\%/50\%} & \footnotesize \textbf{PDS} & \footnotesize \textbf{1/2/4/8\texttimes} & 
             \footnotesize \textbf{FPS} & \footnotesize 1\texttimes \\
            \Xhline{1pt}
        \end{tabular}
    }
    \end{center}
    \vspace{-3mm}
    \label{tab:mvp_comp}
\end{table*}
\begin{figure*}
    \centering
    \vspace{-2.5mm}
    \includegraphics[width=1\linewidth]{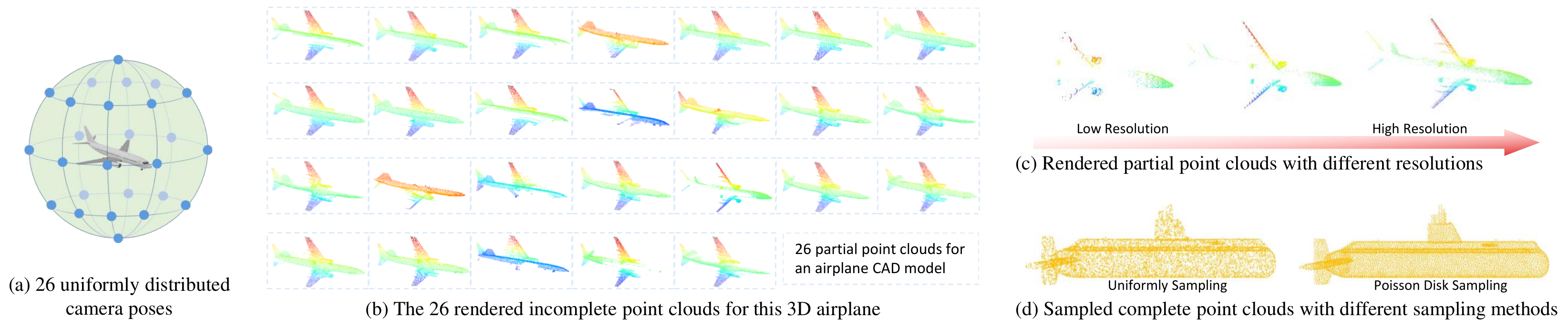}
    \vspace{-7mm}
    \caption{
    \textbf{M}ulti-\textbf{V}iew \textbf{P}artial point cloud dataset. 
    (a) shows an example for 26 uniformly distributed camera poses on a unit sphere. (b) presents the 26 partial point clouds for the airplane from the uniformly distributed virtual cameras. (c) compares the rendered incomplete point clouds with different camera resolutions. (d) shows that Poisson disk sampling generates complete point clouds with a higher quality than uniform sampling.}
    \label{fig:mv_all}
\end{figure*}
\section{Multi-View Partial Point Cloud Dataset}~\label{sec:mvp}
Towards an effort to build a more unified and comprehensive dataset for incomplete point clouds, 
we contribute two high-quality multi-view partial point cloud datasets, MVP and MVP-40 {(based on ShapeNet~\cite{shapenet2015} and ModelNet40~\cite{wu20153d})}, to the community.
We compare MVP and MVP-40 datasets to previous partial point cloud benchmarks in Table~\ref{tab:mvp_comp} (\eg PCN~\cite{yuan2018pcn} and Completion3D~\cite{tchapmi2019topnet}).
Our datasets have many advantages over the other datasets.

\noindent\textbf{Diversity \& Uniform Views.}
First, MVP and MVP-40 datasets consist of diverse partial point clouds.
Instead of rendering partial shapes by using randomly selected camera poses~\cite{yuan2018pcn,tchapmi2019topnet}, we select 26 camera poses that are uniformly distributed on a unit sphere for each CAD model (Fig.~\ref{fig:mv_all} (a)).
Notably, the relative poses between our 26 camera poses are fixed, but the first camera pose is randomly selected, which is equivalent to performing a random rotation to all 26 camera poses.
The major advantages of using uniformly distributed camera views are threefold:
\textbf{1)} MVP datasets have fewer similar rendered partial 3D shapes than the other datasets.
\textbf{2)} The partial point clouds rendered by uniformly distributed camera views can cover most parts of a complete 3D shape.
\textbf{3)} We can generate sufficient incomplete-complete 3D shape pairs with a relatively small number of 3D CAD models.
According to Tatarchenko et. al.~\cite{tatarchenko2019single}, many 3D reconstruction methods rely primarily on shape recognition; they essentially perform shape retrieval from the massive training data.
Hence, using fewer complete shapes during training can better evaluate the capability of generating complete 3D shapes conditioned on the partial observation, rather than naively retrieving a known similar complete shape.
An example of 26 rendered partial point clouds are shown in Fig.~\ref{fig:mv_all} (b).

\noindent\textbf{Large-Scale \& High-Resolution.}
Second, both MVP and MVP-40 consist of over 100,000 high-quality incomplete and complete point clouds.
Poisson Disk Sampling (PDS)~\cite{bridson2007fast,kazhdan2013screened} yields smoother complete point clouds than uniform sampling, making them a better representation of the underlying object CAD models. 
Hence, we employ PDS to sample non-overlapped and uniformly spaced points for complete shapes (Fig.~\ref{fig:mv_all} (d)), which can better evaluate network capabilities of recovering high-quality geometric details.
Previous datasets provide complete shapes with only one resolution.
Unlike those datasets, we create complete point clouds with different resolutions, including 2048(1x), 4096(2x), 8192(4x) and 16384(8x) for precisely evaluating the completion quality at different resolutions.
Specifically, we set missing ratios (\eg $25\%$ or $50\%$) for sampling partial point clouds from high-resolution complete point clouds in the MVP-40 dataset.
As for virtually scanned partial point clouds, previous methods render incomplete point clouds by using small virtual camera resolutions (\eg 160 $\times$ 120), which is much smaller than real depth cameras (\eg both Kinect V2 and Intel RealSense are 1920 $\times$ 1080). 
Consequently, the rendered partial point clouds are unrealistic.
In contrast, we use the resolution 1600 $\times$ 1200 to render partial 3D shapes of high quality (Fig.~\ref{fig:mv_all} (c)) in the MVP dataset.
Note that partial point clouds and the corresponding ground truth have many different points, because they are generated by different sampling processes.

{
\setlength{\tabcolsep}{6.3pt}
\begin{table*}
    \caption{Completion results (CD loss $\times 10^4 \downarrow$) on our MVP dataset (16,384 points). \OM{} outperforms existing methods by convincing margins.}
    \vspace{-4mm}
    \begin{center}
    \small{
        \begin{tabular}{l|cccccccc|cccccccc|c}
            \Xhline{1pt}
                 \multirow{4}{*}{\small Method} & \multirow{4}{*}{\rotatebox{75}{\scriptsize airplane }} & \multirow{4}{*}{\rotatebox{75}{\scriptsize cabinet }} & \multirow{4}{*}{\rotatebox{75}{\scriptsize car }} & \multirow{4}{*}{\rotatebox{75}{\scriptsize chair }} & \multirow{4}{*}{\rotatebox{75}{\scriptsize lamp }} & \multirow{4}{*}{\rotatebox{75}{\scriptsize sofa }} & \multirow{4}{*}{\rotatebox{75}{\scriptsize table }} & \multirow{4}{*}{\rotatebox{75}{\scriptsize watercraft }} & \multirow{4}{*}{\rotatebox{75}{\scriptsize bed }} & \multirow{4}{*}{\rotatebox{75}{\scriptsize bench }} & \multirow{4}{*}{\rotatebox{75}{\scriptsize bookshelf }} & \multirow{4}{*}{\rotatebox{75}{\scriptsize bus }} & \multirow{4}{*}{\rotatebox{75}{\scriptsize guitar }} & \multirow{4}{*}{\rotatebox{75}{\scriptsize motorbike }} & \multirow{4}{*}{\rotatebox{75}{\scriptsize pistol }} & \multirow{4}{*}{\rotatebox{75}{\scriptsize skateboard }} & \multirow{4}{*}{\small Avg.} \\
                  & & & & & & & & & & & & & & & & & \\
                  & & & & & & & & & & & & & & & & & \\
                  & & & & & & & & & & & & & & & & & \\
             \hline\hline
            \scriptsize PCN~\cite{yuan2018pcn} & \scriptsize 2.95 & \scriptsize 4.13 & \scriptsize 3.04 & \scriptsize 7.07 & \scriptsize 14.93 & \scriptsize 5.56 & \scriptsize 7.06 & \scriptsize 6.08 & \scriptsize 12.72 & \scriptsize 5.73 & \scriptsize 6.91 & \scriptsize 2.46 & \scriptsize 1.02 & \scriptsize 3.53 & \scriptsize 3.28 & \scriptsize 2.99 & \scriptsize 6.02 \\
             
            \scriptsize TopNet~\cite{tchapmi2019topnet} & \scriptsize 2.72 & \scriptsize 4.25 & \scriptsize 3.40 & \scriptsize 7.95 & \scriptsize 17.01 & \scriptsize 6.04 & \scriptsize 7.42 & \scriptsize 6.04 & \scriptsize 11.60 & \scriptsize 5.62 & \scriptsize 8.22 & \scriptsize 2.37 & \scriptsize 1.33 & \scriptsize 3.90 & \scriptsize 3.97 & \scriptsize 2.09 & \scriptsize 6.36 \\
             
            \scriptsize MSN~\cite{liu2020morphing} & \scriptsize 2.07 & \scriptsize 3.82 & \scriptsize 2.76 & \scriptsize 6.21 & \scriptsize 12.72 & \scriptsize 4.74 & \scriptsize 5.32 & \scriptsize 4.80 & \scriptsize 9.93 & \scriptsize 3.89 & \scriptsize 5.85 & \scriptsize 2.12 & \scriptsize 0.69 & \scriptsize 2.48 & \scriptsize 2.91 & \scriptsize 1.58 & \scriptsize 4.90 \\
             
            \scriptsize Wang et. al.~\cite{wang2020cascaded} & \scriptsize 1.59 & \scriptsize 3.64 & \scriptsize 2.60 & \scriptsize 5.24 & \scriptsize 9.02 & \scriptsize 4.42 & \scriptsize 5.45 & \scriptsize 4.26 & \scriptsize 9.56 & \scriptsize 3.67 & \scriptsize 5.34 & \scriptsize 2.23 & \scriptsize 0.79 & \scriptsize 2.23 & \scriptsize 2.86 & \scriptsize 2.13 & \scriptsize 4.30 \\
             
            \scriptsize ECG~\cite{pan2020ecg} & \scriptsize 1.41 & \scriptsize 3.44 & \scriptsize 2.36 & \scriptsize 4.58 & \scriptsize 6.95 & \scriptsize 3.81 & \scriptsize 4.27 & \scriptsize 3.38 & \scriptsize 7.46 & \scriptsize 3.10 & \scriptsize 4.82 & \scriptsize 1.99 & \scriptsize 0.59 & \scriptsize 2.05 & \scriptsize 2.31 & \scriptsize 1.66 & \scriptsize 3.58 \\
             
            \scriptsize GRNet~\cite{xie2020grnet} & \scriptsize 1.61 & \scriptsize 4.66 & \scriptsize 3.10 & \scriptsize 4.72 & \scriptsize 5.66 & \scriptsize 4.61 & \scriptsize 4.85 & \scriptsize 3.53 & \scriptsize 7.82 & \scriptsize 2.96 & \scriptsize 4.58 & \scriptsize 2.97 & \scriptsize 1.28 & \scriptsize 2.24 & \scriptsize 2.11 & \scriptsize 1.61 & \scriptsize 3.87 \\
            
            \scriptsize NSFA~\cite{zhang2020detail} & \scriptsize 1.51 & \scriptsize 4.24 & \scriptsize 2.75 & \scriptsize 4.68 & \scriptsize 6.04 & \scriptsize 4.29 & \scriptsize 4.84 & \scriptsize 3.02 & \scriptsize 7.93 & \scriptsize 3.87 & \scriptsize 5.99 & \scriptsize 2.21 & \scriptsize 0.78 & \scriptsize 1.73 & \scriptsize 2.04 & \scriptsize 2.14 & \scriptsize 3.77 \\
            \hline
            
            \scriptsize \OM{} (Ours) & \scriptsize \textbf{1.15} & \scriptsize \textbf{3.20} & \scriptsize \textbf{2.14} & \scriptsize \textbf{3.58} & \scriptsize \textbf{5.57} & \scriptsize \textbf{3.58} & \scriptsize \textbf{4.17} & \scriptsize \textbf{2.47} & \scriptsize \textbf{6.90} & \scriptsize \textbf{2.76} & \scriptsize \textbf{3.45} & \scriptsize \textbf{1.78} & \scriptsize \textbf{0.59} & \scriptsize \textbf{1.52} & \scriptsize \textbf{1.83} & \scriptsize \textbf{1.57} & \scriptsize \textbf{3.06} \\

            \Xhline{1pt}
            
        \end{tabular}
    }
    \end{center}

    \label{tab:mvpc_CD_16384}
    \vspace{-5mm}
\end{table*}
}
{
\setlength{\tabcolsep}{5.95pt}
\begin{table*}
    \vspace{1.5mm}
    \caption{Completion results (F-Score@1\% $\uparrow$) on our 
    MVP dataset (16,384 points). 
    }
    \vspace{-4mm}
    \begin{center}
        \small{
            \begin{tabular}{l|cccccccc|cccccccc|c}
            \Xhline{1pt}
            \multirow{4}{*}{\small Method} & \multirow{4}{*}{\rotatebox{75}{\scriptsize airplane }} & \multirow{4}{*}{\rotatebox{75}{\scriptsize cabinet }} & \multirow{4}{*}{\rotatebox{75}{\scriptsize car }} & \multirow{4}{*}{\rotatebox{75}{\scriptsize chair }} & \multirow{4}{*}{\rotatebox{75}{\scriptsize lamp }} & \multirow{4}{*}{\rotatebox{75}{\scriptsize sofa }} & \multirow{4}{*}{\rotatebox{75}{\scriptsize table }} & \multirow{4}{*}{\rotatebox{75}{\scriptsize watercraft }} & \multirow{4}{*}{\rotatebox{75}{\scriptsize bed }} & \multirow{4}{*}{\rotatebox{75}{\scriptsize bench }} & \multirow{4}{*}{\rotatebox{75}{\scriptsize bookshelf }} & \multirow{4}{*}{\rotatebox{75}{\scriptsize bus }} & \multirow{4}{*}{\rotatebox{75}{\scriptsize guitar }} & \multirow{4}{*}{\rotatebox{75}{\scriptsize motorbike }} & \multirow{4}{*}{\rotatebox{75}{\scriptsize pistol }} & \multirow{4}{*}{\rotatebox{75}{\scriptsize skateboard }} & \multirow{4}{*}{\small Avg.} \\
                  & & & & & & & & & & & & & & & & & \\
                  & & & & & & & & & & & & & & & & & \\
                  & & & & & & & & & & & & & & & & & \\
             \hline\hline
            \scriptsize PCN~\cite{yuan2018pcn} & \fontsize{6}{6}\selectfont 0.816 & \fontsize{6}{6}\selectfont 0.614 & \fontsize{6}{6}\selectfont 0.686 & \fontsize{6}{6}\selectfont 0.517 & \fontsize{6}{6}\selectfont 0.455 & \fontsize{6}{6}\selectfont 0.552 & \fontsize{6}{6}\selectfont 0.646 & \fontsize{6}{6}\selectfont 0.628 & \fontsize{6}{6}\selectfont 0.452 & \fontsize{6}{6}\selectfont 0.694 & \fontsize{6}{6}\selectfont 0.546 & \fontsize{6}{6}\selectfont 0.779 & \fontsize{6}{6}\selectfont 0.906 & \fontsize{6}{6}\selectfont 0.665 & \fontsize{6}{6}\selectfont 0.774 & \fontsize{6}{6}\selectfont 0.861 & \fontsize{6}{6}\selectfont 0.638 \\
             
            \scriptsize TopNet~\cite{tchapmi2019topnet} & \fontsize{6}{6}\selectfont 0.789 & \fontsize{6}{6}\selectfont 0.621 & \fontsize{6}{6}\selectfont 0.612 & \fontsize{6}{6}\selectfont 0.443 & \fontsize{6}{6}\selectfont 0.387 & \fontsize{6}{6}\selectfont 0.506 & \fontsize{6}{6}\selectfont 0.639 & \fontsize{6}{6}\selectfont 0.609 & \fontsize{6}{6}\selectfont 0.405 & \fontsize{6}{6}\selectfont 0.680 & \fontsize{6}{6}\selectfont 0.524 & \fontsize{6}{6}\selectfont 0.766 & \fontsize{6}{6}\selectfont 0.868 & \fontsize{6}{6}\selectfont 0.619 & \fontsize{6}{6}\selectfont 0.726 & \fontsize{6}{6}\selectfont 0.837 & \fontsize{6}{6}\selectfont 0.601 \\
             
            \scriptsize MSN~\cite{liu2020morphing} & \fontsize{6}{6}\selectfont 0.879 & \fontsize{6}{6}\selectfont 0.692 & \fontsize{6}{6}\selectfont 0.693 & \fontsize{6}{6}\selectfont 0.599 & \fontsize{6}{6}\selectfont 0.604 & \fontsize{6}{6}\selectfont 0.627 & \fontsize{6}{6}\selectfont 0.730 & \fontsize{6}{6}\selectfont 0.696 & \fontsize{6}{6}\selectfont 0.569 & \fontsize{6}{6}\selectfont 0.797 & \fontsize{6}{6}\selectfont 0.637 & \fontsize{6}{6}\selectfont 0.806 & \fontsize{6}{6}\selectfont 0.935 & \fontsize{6}{6}\selectfont 0.728 & \fontsize{6}{6}\selectfont 0.809 & \fontsize{6}{6}\selectfont 0.885 & \fontsize{6}{6}\selectfont 0.710 \\
             
            \scriptsize Wang et. al.~\cite{wang2020cascaded} & \fontsize{6}{6}\selectfont 0.898 & \fontsize{6}{6}\selectfont 0.688 & \fontsize{6}{6}\selectfont 0.725 & \fontsize{6}{6}\selectfont 0.670 & \fontsize{6}{6}\selectfont 0.681 & \fontsize{6}{6}\selectfont 0.641 & \fontsize{6}{6}\selectfont 0.748 & \fontsize{6}{6}\selectfont 0.742 & \fontsize{6}{6}\selectfont 0.600 & \fontsize{6}{6}\selectfont 0.797 & \fontsize{6}{6}\selectfont 0.659 & \fontsize{6}{6}\selectfont 0.802 & \fontsize{6}{6}\selectfont 0.931 & \fontsize{6}{6}\selectfont 0.772 & \fontsize{6}{6}\selectfont 0.843 & \fontsize{6}{6}\selectfont 0.902 & \fontsize{6}{6}\selectfont 0.740 \\
             
            \scriptsize ECG~\cite{pan2020ecg} & \fontsize{6}{6}\selectfont 0.906 & \fontsize{6}{6}\selectfont 0.680 & \fontsize{6}{6}\selectfont 0.716 & \fontsize{6}{6}\selectfont 0.683 & \fontsize{6}{6}\selectfont 0.734 & \fontsize{6}{6}\selectfont 0.651 & \fontsize{6}{6}\selectfont 0.766 & \fontsize{6}{6}\selectfont 0.753 & \fontsize{6}{6}\selectfont 0.640 & \fontsize{6}{6}\selectfont 0.822 & \fontsize{6}{6}\selectfont 0.706 & \fontsize{6}{6}\selectfont 0.804 & \fontsize{6}{6}\selectfont 0.945 & \fontsize{6}{6}\selectfont 0.780 & \fontsize{6}{6}\selectfont 0.835 & \fontsize{6}{6}\selectfont 0.897 & \fontsize{6}{6}\selectfont 0.753 \\
             
            \scriptsize GRNet~\cite{xie2020grnet} & \fontsize{6}{6}\selectfont 0.853 & \fontsize{6}{6}\selectfont 0.578 & \fontsize{6}{6}\selectfont 0.646 & \fontsize{6}{6}\selectfont 0.635 & \fontsize{6}{6}\selectfont 0.710 & \fontsize{6}{6}\selectfont 0.580 & \fontsize{6}{6}\selectfont 0.690 & \fontsize{6}{6}\selectfont 0.723 & \fontsize{6}{6}\selectfont 0.586 & \fontsize{6}{6}\selectfont 0.765 & \fontsize{6}{6}\selectfont 0.635 & \fontsize{6}{6}\selectfont 0.682 & \fontsize{6}{6}\selectfont 0.865 & \fontsize{6}{6}\selectfont 0.736 & \fontsize{6}{6}\selectfont 0.787 & \fontsize{6}{6}\selectfont 0.850 & \fontsize{6}{6}\selectfont 0.692 \\
            
            \scriptsize NSFA~\cite{zhang2020detail} & \fontsize{6}{6}\selectfont 0.903 & \fontsize{6}{6}\selectfont 0.694 & \fontsize{6}{6}\selectfont 0.721 & \fontsize{6}{6}\selectfont 0.737 & \fontsize{6}{6}\selectfont 0.783 & \fontsize{6}{6}\selectfont \textbf{0.705} & \fontsize{6}{6}\selectfont \textbf{0.817} & \fontsize{6}{6}\selectfont 0.799 & \fontsize{6}{6}\selectfont \textbf{0.687} & \fontsize{6}{6}\selectfont 0.845 & \fontsize{6}{6}\selectfont 0.747 & \fontsize{6}{6}\selectfont 0.815 & \fontsize{6}{6}\selectfont 0.932 & \fontsize{6}{6}\selectfont 0.815 & \fontsize{6}{6}\selectfont 0.858 & \fontsize{6}{6}\selectfont 0.894 & \fontsize{6}{6}\selectfont 0.783 \\
            \hline
            
            \scriptsize \OM{} (Ours) & \fontsize{6}{6}\selectfont \textbf{0.928} & \fontsize{6}{6}\selectfont \textbf{0.721} & \fontsize{6}{6}\selectfont \textbf{0.756} & \fontsize{6}{6}\selectfont \textbf{0.743} & \fontsize{6}{6}\selectfont \textbf{0.789} & \fontsize{6}{6}\selectfont 0.696 & \fontsize{6}{6}\selectfont 0.813 & \fontsize{6}{6}\selectfont \textbf{0.800} & \fontsize{6}{6}\selectfont 0.674 & \fontsize{6}{6}\selectfont \textbf{0.863} & \fontsize{6}{6}\selectfont \textbf{0.755} & \fontsize{6}{6}\selectfont \textbf{0.832} & \fontsize{6}{6}\selectfont \textbf{0.960} & \fontsize{6}{6}\selectfont \textbf{0.834} & \fontsize{6}{6}\selectfont \textbf{0.887} & \fontsize{6}{6}\selectfont \textbf{0.930} & \fontsize{6}{6}\selectfont \textbf{0.796} \\
            \Xhline{1pt}
            
        \end{tabular}
        }
    \end{center}

    \label{tab:mvpc_F1_16384}
    \vspace{-5mm}
\end{table*}
}

\noindent\textbf{Rich Categories.}
Third, the MVP dataset consists of 16 shape categories of partial and complete shapes for training and testing.
Besides the 8 categories (airplane, cabinet, car, chair, lamp, sofa, table and watercraft) included in previous datasets~\cite{yuan2018pcn,tchapmi2019topnet}, we add another 8 categories (bed, bench, bookshelf, bus, guitar, motorbike, pistol and skateboard).
Moreover, the MVP-40 dataset consists of 40 shape categories, which makes it more challenging to train and evaluate networks for completion and perception.

To sum up, MVP and MVP-40 datasets consist of a large number of high-quality synthetic partial scans for 3D CAD models, and their incompleteness are mainly caused by self-occlusion.
Besides 3D shape completion, our datasets can be used in many other partial point cloud tasks, such as classification, registration and keypoints extraction.
Compared to previous partial point cloud datasets, MVP and MVP-40 have many favorable properties.

\section{Experiments} \label{sec:Exp}
\noindent\textbf{Evaluation Metrics.}
In line with previous methods~\cite{tchapmi2019topnet,xie2020grnet,zhang2020detail}, we evaluate the reconstruction accuracy by computing the Chamfer Distance (Eq.~\eqref{eq:cd_loss}) between the predicted complete shapes $\mathbf{Y}^{\prime}$ and the ground truth shapes $\mathbf{Y}$.
Based on the insight that CD can be misleading due to its sensitivity to outliers~\cite{tatarchenko2019single}, we also use F-score~\cite{knapitsch2017tanks} to evaluate the distance between object surfaces, which is defined as the harmonic mean between precision and recall.
As for classification tasks, we use the two metrics, overall accuracy for all evaluated instances (Acc.) and average accuracy among all categories (Avg.).

\noindent\textbf{Implementation Details.}
Our networks are implemented using PyTorch.
We train our models using the Adam optimizer~\cite{kingma2014adam} with initial learning rate 1e$^{-4}$ (decayed by 0.7 every 40 epochs) and batch size 32 by NVIDIA TITAN Xp GPU.
Note that \OM{} does not use any symmetry tricks, such as reflection symmetry or mirror operations.
As for classification, we evaluate different completion results by using the same PointNet and DGCNN networks that are pretrained on complete point clouds.

\setlength{\tabcolsep}{5pt}
\begin{table}
    \caption{Completion results (CD loss $\times 10^4$) with various resolutions. 
    }
    \vspace{-4mm}
    \begin{center}
    \small{
        \begin{tabular}{l|cc|cc|cc|cc}
            \Xhline{1pt}
            \multirow{2}{*}{\small\# Points} & \multicolumn{2}{c|}{\scriptsize{2,048}} & \multicolumn{2}{c|}{\scriptsize{4,096}} & \multicolumn{2}{c|}{\scriptsize{8,192}} & \multicolumn{2}{c}{\scriptsize{16,384}} \\
            \cline{2-9}
             & \scriptsize CD & \scriptsize F1 & \scriptsize CD & \scriptsize F1 & \scriptsize CD & \scriptsize F1 & \scriptsize CD & \scriptsize F1  \\
            \hline\hline
            \scriptsize PCN~\cite{yuan2018pcn} & \fontsize{6}{6}\selectfont 9.77 & \fontsize{6}{6}\selectfont 0.320 & \fontsize{6}{6}\selectfont 7.96 & \fontsize{6}{6}\selectfont 0.458 & \fontsize{6}{6}\selectfont 6.99 & \fontsize{6}{6}\selectfont 0.563 & \fontsize{6}{6}\selectfont 6.02 & \fontsize{6}{6}\selectfont 0.638\\
            \scriptsize TopNet~\cite{tchapmi2019topnet} & \fontsize{6}{6}\selectfont 10.11 & \fontsize{6}{6}\selectfont 0.308 & \fontsize{6}{6}\selectfont 8.20 & \fontsize{6}{6}\selectfont 0.440 & \fontsize{6}{6}\selectfont 7.00 & \fontsize{6}{6}\selectfont 0.533 & \fontsize{6}{6}\selectfont 6.36 & \fontsize{6}{6}\selectfont 0.601 \\
            \scriptsize MSN~\cite{liu2020morphing} & \fontsize{6}{6}\selectfont 7.90 & \fontsize{6}{6}\selectfont 0.432 & \fontsize{6}{6}\selectfont 6.17 & \fontsize{6}{6}\selectfont 0.585 & \fontsize{6}{6}\selectfont 5.42 & \fontsize{6}{6}\selectfont 0.659 & \fontsize{6}{6}\selectfont 4.90 & \fontsize{6}{6}\selectfont 0.710 \\
            \scriptsize Wang et. al.~\cite{wang2020cascaded} & \fontsize{6}{6}\selectfont 7.25 & \fontsize{6}{6}\selectfont 0.434 & \fontsize{6}{6}\selectfont 5.83 & \fontsize{6}{6}\selectfont 0.569 & \fontsize{6}{6}\selectfont 4.90 & \fontsize{6}{6}\selectfont 0.680 & \fontsize{6}{6}\selectfont 4.30 & \fontsize{6}{6}\selectfont 0.740 \\
            \scriptsize ECG~\cite{pan2020ecg} & \fontsize{6}{6}\selectfont 6.64 & \fontsize{6}{6}\selectfont 0.476 & \fontsize{6}{6}\selectfont 5.41 & \fontsize{6}{6}\selectfont 0.585 & \fontsize{6}{6}\selectfont 4.18 & \fontsize{6}{6}\selectfont 0.690 & \fontsize{6}{6}\selectfont 3.58 & \fontsize{6}{6}\selectfont 0.753 \\
            \hline
            \scriptsize \OM{} (Ours) & \fontsize{6}{6}\selectfont \textbf{5.96} & \fontsize{6}{6}\selectfont \textbf{0.499} & \fontsize{6}{6}\selectfont \textbf{4.70} & \fontsize{6}{6}\selectfont \textbf{0.636} & \fontsize{6}{6}\selectfont \textbf{3.64} & \fontsize{6}{6}\selectfont \textbf{0.727} & \fontsize{6}{6}\selectfont \textbf{3.12} & \fontsize{6}{6}\selectfont \textbf{0.791}\\
            \Xhline{1pt}
        \end{tabular}
    }
    \end{center}
    \vspace{-4mm}
    \label{tab:res}
\end{table}
\subsection{Completion and Classification on MVP Dataset}

\noindent\textbf{Completion Evaluation.}
As introduced in Sec.~\ref{sec:mvp}, MVP dataset consists of 16 categories of high-quality partial/complete point clouds that are generated by CAD models selected from the ShapeNet~\cite{shapenet2015} dataset.
We split our models into a training set (62,400 shape pairs) and a test set (41,600 shape pairs).
Similarly, MVP-40 dataset consists of a training set (41,600 shape pairs) and a test set (64,168 shape pairs) from 40 categories.
Notably, none of the complete shapes in our test set are included in our training set.
To achieve a fair comparison, we train all methods using the same training strategy on MVP dataset.
The evaluated CD loss and F-score for all evaluated methods (16,384 points) are reported in Table~\ref{tab:mvpc_CD_16384} and Table~\ref{tab:mvpc_F1_16384}, respectively.
\OM{} outperforms all existing competitive methods in terms of CD and F-score@1\%.
Moreover, \OM{} can generate complete point clouds with various resolutions ($N=$ 2048, 4096, 8192 and 16384). 
We compare our methods with existing approaches that support multi-resolution completion in Table~\ref{tab:res}, and \OM{} outperforms all the other methods.
\begin{figure*}
    \centering
    \includegraphics[width=1\linewidth]{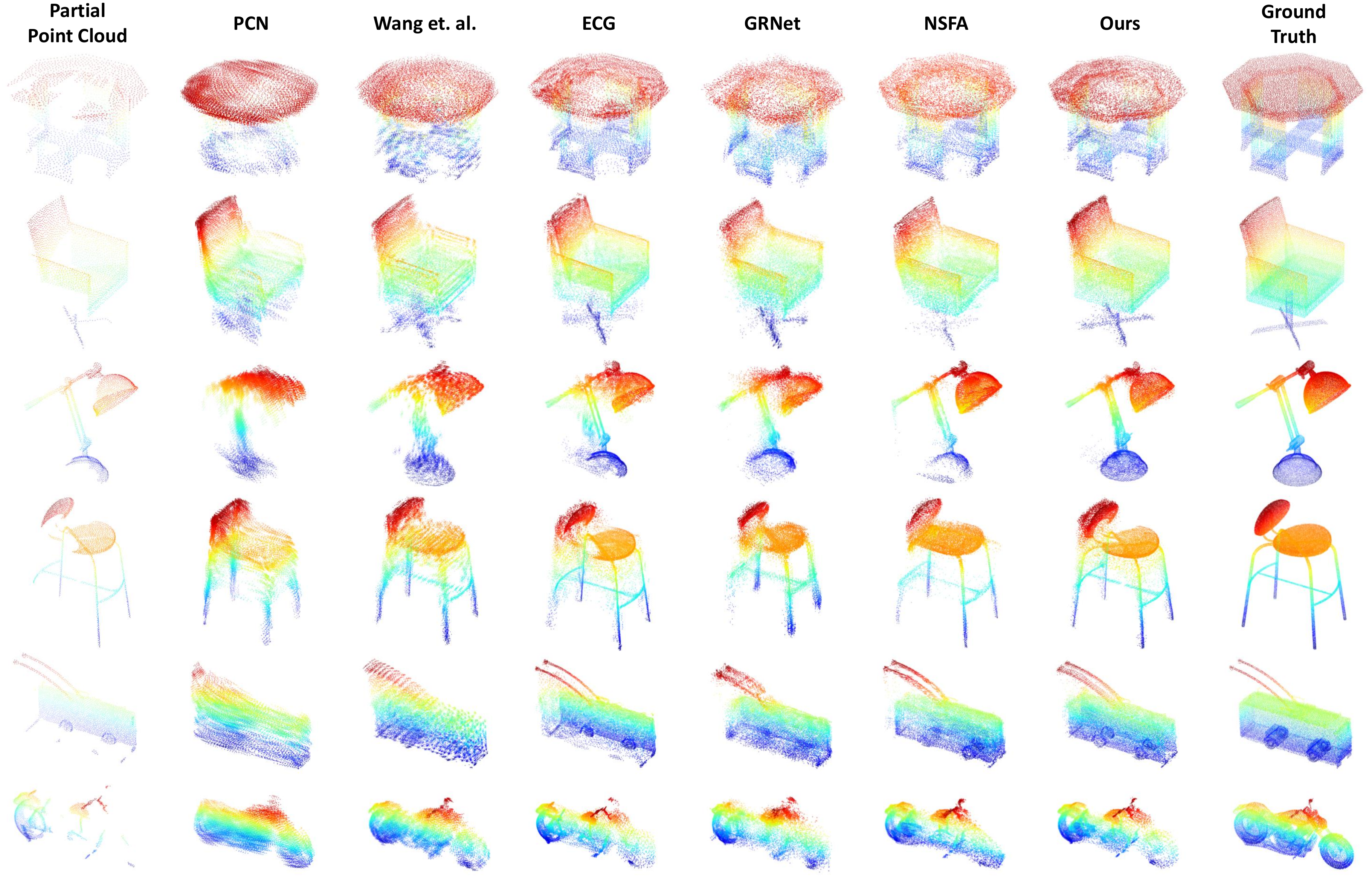}
    \vspace{-7mm}
    \caption{\textbf{Qualitative completion results (16,384 points) on the MVP dataset by different methods.} \OM{} can generate better complete point clouds than the other methods by learning geometrical symmetries.}
    \vspace{-3mm}
    \label{fig:qua_all}
\end{figure*}

The qualitative comparison results are shown in Fig.~\ref{fig:qua_all}.
The proposed \OM{} can generate better complete shapes with fine details than the other methods.
In particular, we can clearly observe the learned relational structures in our complete shapes.
For example, the missing legs of the chairs (the second row and the fourth row in Fig.~\ref{fig:qua_all}) are recovered based on the observed legs with the learned shape symmetry.
In the third row of Fig.~\ref{fig:qua_all}, we reconstruct the incomplete lamp base with a smooth round bowl shape, which makes it a more plausible completion than the others.
The partially observed motorbike in the last row does not contain its front wheel, and \OM{} reconstructs a complete wheel by learning the observed back wheel.
Consequently, \OM{} can effectively reconstruct complete shapes by learning structural relations, including geometrical symmetries, regular arrangements and surface smoothness, from the incomplete point cloud.

{
\setlength{\tabcolsep}{7.3pt}
\begin{table}[]
    \caption{Ablation studies for \OM{} (2,048 points).}
    \vspace{-4mm}
    \begin{center}
        \small{
        \begin{tabular}{ccc|cc}
            \Xhline{1pt}
            \makecell{\footnotesize{Point} \\ \footnotesize{Self-Attention}} & \makecell{\footnotesize{Dual-Path} \\ \footnotesize{Architecture}} & \makecell{\footnotesize{Kernel} \\ \footnotesize{Selection}} & \scriptsize CD & \scriptsize F1 \\
            \hline\hline
             & & & \footnotesize 6.64 & \footnotesize 0.476  \\
             & \checkmark & & \footnotesize 6.43 & \footnotesize 0.488  \\
            \checkmark & & & \footnotesize 6.35 & \footnotesize 0.484  \\
             & \checkmark & \checkmark & \footnotesize 6.35 & \footnotesize 0.490  \\
            \checkmark & \checkmark & & \footnotesize 6.15 & \footnotesize 0.492 \\
            \checkmark & \checkmark & \checkmark & \footnotesize 5.96 & \footnotesize 0.499 \\
            \Xhline{1pt}
        \end{tabular}
        }
    \end{center}
    \vspace{-3mm}
    \label{tab:ablation}
\end{table}
}

\noindent\textbf{Ablation Study.}
The ablation studies for all our proposed modules, Point Self-Attention Kernel (PSA), Dual-path Architecture and Kernel Selection (two-branch PSK), are presented in Table~\ref{tab:ablation}.
We use ECG~\cite{pan2020ecg} as our baseline model and evaluate the completion results with 2048 points.
By adding the proposed modules, better completion results can be achieved, which validates their effectiveness.

{
\setlength{\tabcolsep}{6.6pt}
\begin{table}[t]
    \caption{Classification results on MVP dataset (2,048 points).}
    \vspace{-4mm}
    \begin{center}
    \small{
        \begin{tabular}{l|cc|cc}
            \Xhline{1pt}
            \multirow{2}{*}{{\small{Method}}} & \multicolumn{2}{c|}{{\footnotesize{PointNet~\cite{qi2017pointnet}}}} & \multicolumn{2}{c}{{\footnotesize{DGCNN~\cite{dgcnn}}}} \\
            \cline{2-5}
             & \scriptsize{Acc. (\%)} & \scriptsize{Avg. (\%)} & \scriptsize{Acc. (\%)} & \scriptsize{Avg. (\%)}  \\
             \hline\hline
            \footnotesize{Partial} & \scriptsize{70.5} & \scriptsize{72.5} & \scriptsize{68.6} & \scriptsize{69.8} \\
            \; \scriptsize{+ PCN~\cite{yuan2018pcn}} & \scriptsize{86.2} & \scriptsize{83.7} & \scriptsize{81.0} & \scriptsize{79.7} \\
            \; \footnotesize{+ TopNet~\cite{tchapmi2019topnet}} & \scriptsize{84.9} & \scriptsize{82.1} & \scriptsize{74.8} & \scriptsize{71.2} \\
            \; \footnotesize{+ ECG~\cite{pan2020ecg}} & \scriptsize{86.5} & \scriptsize{84.0} & \scriptsize{80.7} & \scriptsize{79.4} \\
            \; \footnotesize{+ VRCNet (Ours)} & \textbf{\scriptsize{87.2}} & \textbf{\scriptsize{85.3}} & \textbf{\scriptsize{82.0}} & \textbf{\scriptsize{81.1}} \\
            \hline
            \footnotesize{Complete} & {\scriptsize{90.9}} & {\scriptsize{90.1}} & {\scriptsize{91.9}} & {\scriptsize{91.4}} \\
            \Xhline{1pt}
            
        \end{tabular}
    }    
    \end{center}
    \vspace{-4mm}
    \label{tab:mvp_cf}
\end{table}
}

\noindent\textbf{Classification Evaluation.}
In Table~\ref{tab:mvp_cf}, we evaluate the classification performance by using overall accuracy (Acc.) and average category accuracy (Avg.) on MVP dataset.
We use two networks, PointNet~\cite{qi2017pointnet} and DGCNN~\cite{dgcnn}, which are trained with complete point clouds from the training set of MVP.
Afterwards, we evaluate the trained models for point clouds from the test set of MVP with different settings, including ``Partial'' (partial point clouds), ``+PCN'' (completion results by PCN), ``+\OM{}'' (completion results by \OM{}) and ``Complete'' (ground truth complete point clouds).
Comparing with complete point clouds, clear performance drops (\ie $> 20\%$) of both PointNet and DGCNN can be observed for classifying partial point clouds.
Furthermore, completion results by various methods improve the classification accuracy.
In particular, completion results by \OM{} can be more accurately classified than those by the other methods, which further validates better completion capability of \OM{}.

\setlength{\tabcolsep}{5pt}
\begin{table}
    \caption{Completion results (CD $\times 10^4$) on Completion3D.
    }
    \vspace{-4mm}
    \begin{center}
    \small{
        \begin{tabular}{l|cccccccc|c}
            \Xhline{1pt}
            \multirow{3}{*}{{\small Method}} & \multirow{3}{*}{\rotatebox{75}{\fontsize{5}{5}\selectfont airplane }} & \multirow{3}{*}{\rotatebox{75}{\fontsize{5}{5}\selectfont cabinet }} & \multirow{3}{*}{\rotatebox{75}{\fontsize{5}{5}\selectfont car }} & \multirow{3}{*}{\rotatebox{75}{\fontsize{5}{5}\selectfont chair }} & \multirow{3}{*}{\rotatebox{75}{\fontsize{5}{5}\selectfont lamp }} & \multirow{3}{*}{\rotatebox{75}{\fontsize{5}{5}\selectfont sofa }} & \multirow{3}{*}{\rotatebox{75}{\fontsize{5}{5}\selectfont table }} & \multirow{3}{*}{\rotatebox{75}{\fontsize{5}{5}\selectfont watercraft }} & \multirow{3}{*}{\small{Avg.}} \\
            & & & & & & & & \\
            & & & & & & & & \\
             \hline\hline
            \fontsize{5}{5}\selectfont AtlasNet~\cite{groueix2018atlasnet} & \fontsize{5}{5}\selectfont 10.36 & \fontsize{5}{5}\selectfont 23.40 & \fontsize{5}{5}\selectfont 13.40 & \fontsize{5}{5}\selectfont 24.16 & \fontsize{5}{5}\selectfont 20.24 & \fontsize{5}{5}\selectfont 20.82 & \fontsize{5}{5}\selectfont 17.52 & \fontsize{5}{5}\selectfont 11.62 & \fontsize{5}{5}\selectfont 17.77 \\ 
            
            \fontsize{5}{5}\selectfont PCN~\cite{yuan2018pcn} & \fontsize{5}{5}\selectfont 9.79 & \fontsize{5}{5}\selectfont 22.70 & \fontsize{5}{5}\selectfont 12.43 & \fontsize{5}{5}\selectfont 25.14 & \fontsize{5}{5}\selectfont 22.72 & \fontsize{5}{5}\selectfont 20.26 & \fontsize{5}{5}\selectfont 20.27 & \fontsize{5}{5}\selectfont 11.73 & \fontsize{5}{5}\selectfont 18.22 \\ 
             
            \fontsize{5}{5}\selectfont TopNet~\cite{tchapmi2019topnet} & \fontsize{5}{5}\selectfont 7.32 & \fontsize{5}{5}\selectfont 18.77 & \fontsize{5}{5}\selectfont 12.88 & \fontsize{5}{5}\selectfont 19.82 & \fontsize{5}{5}\selectfont 14.60 & \fontsize{5}{5}\selectfont 16.29 & \fontsize{5}{5}\selectfont 14.89 & \fontsize{5}{5}\selectfont 8.82 & \fontsize{5}{5}\selectfont 14.25 \\ 
             
            \fontsize{5}{5}\selectfont GRNet~\cite{xie2020grnet} & \fontsize{5}{5}\selectfont 6.13 & \fontsize{5}{5}\selectfont 16.90 & \fontsize{5}{5}\selectfont 8.27 & \fontsize{5}{5}\selectfont 12.23 & \fontsize{5}{5}\selectfont 10.22 & \fontsize{5}{5}\selectfont 14.93 & \fontsize{5}{5}\selectfont 10.08 & \fontsize{5}{5}\selectfont 5.86 & \fontsize{5}{5}\selectfont 10.64 \\
            
            \hline
            
            \fontsize{5}{5}\selectfont \OM{} (Ours) & \fontsize{5}{5}\selectfont \textbf{3.94} & \fontsize{5}{5}\selectfont \textbf{10.93} & \fontsize{5}{5}\selectfont \textbf{6.44} & \fontsize{5}{5}\selectfont \textbf{9.32} & \fontsize{5}{5}\selectfont \textbf{8.32} & \fontsize{5}{5}\selectfont \textbf{11.35} & \fontsize{5}{5}\selectfont \textbf{8.60} & \fontsize{5}{5}\selectfont \textbf{5.78} & \fontsize{5}{5}\selectfont \textbf{8.12} \\
            \Xhline{1pt}
            
        \end{tabular}
    }    
    \end{center}
    \vspace{-4mm}
    \label{tab:C3D_CD_2048}
\end{table}
\begin{figure*}
    \centering
    \includegraphics[width=1\linewidth]{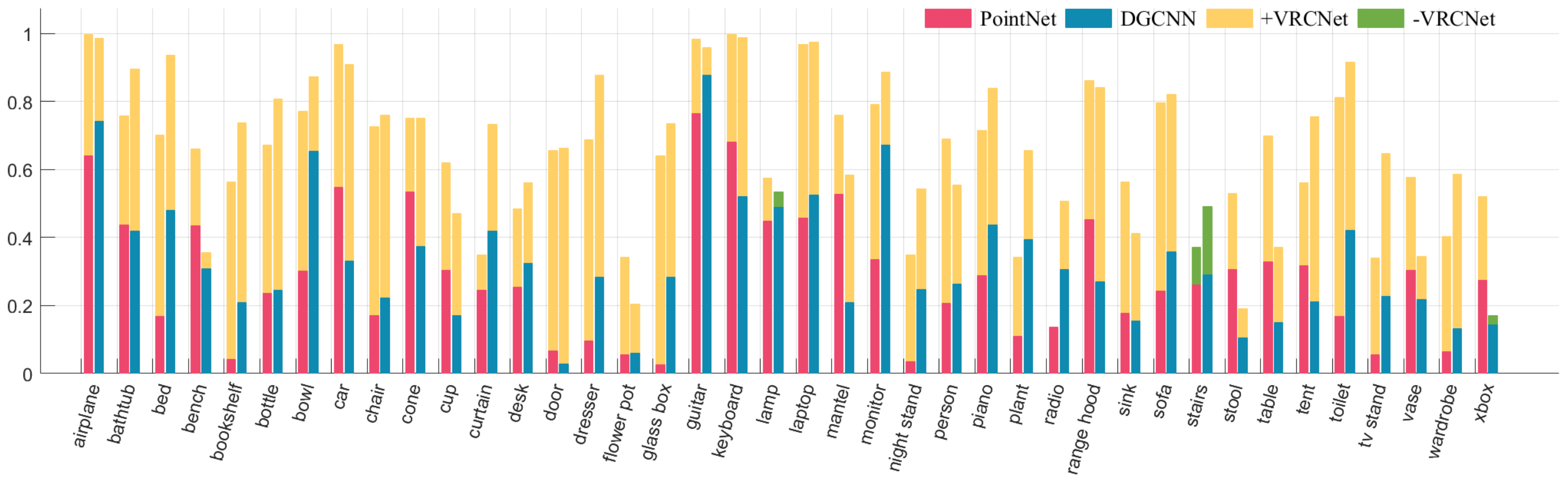}
    \vspace{-7mm}
    \caption{\textbf{Per-category classification results (Acc. \%) on MVP-40 dataset (50\% missing).} 
    PointNet~\cite{qi2017pointnet} ({\color[RGB]{238,71,110}\textbf{red}} bars) and DGCNN~\cite{dgcnn} ({\color[RGB]{15,138,177}\textbf{blue}} bars) can provide much higher classification accuracy ({\color[RGB]{255,208,102}\textbf{yellow}} bar) with the help of point cloud completion by \OM{} than those directly on partial point clouds for most categories, although accuracy decrements ({\color[RGB]{112, 173, 71}\textbf{green}} bars) are observed in few categories.
    }
    \label{fig:all_cat}
\end{figure*}

\subsection{Completion on Completion3D Dataset}
The Completion3D benchmark is an online platform for evaluating 3D shape completion approaches.
Following their instructions, we train \OM{} using their prepared training data and upload our best completion results (2,048 points). 
As reported in the online leaderboard{\href{https://completion3d.stanford.edu/results}{$^1$}}, also shown in Table~\ref{tab:C3D_CD_2048}, \OM{} significantly outperforms SoTA methods and is ranked first on the Completion3D benchmark.

\subsection{Completion and Classification on MVP-40 Dataset}
ModelNet40~\cite{wu20153d} is a large-scale 3D CAD Dataset with 40 common object categories, based on which we construct the MVP-40 dataset (see Table~\ref{tab:mvp_comp}).
For each category, we randomly select 40 different models (totally 1,600 models) for training,
and we use the same 2,468 models with PointNet~\cite{qi2017pointnet} to construct our testing set.
In line with MVP, we randomly select 26 uniformly distributed virtual camera poses for sampling diverse partial point clouds, and the groundtruth complete point clouds are generated by PDS. 
Unlike MVP, we define a specific missing ratio to generate incomplete point clouds by discarding those points that are far from current camera viewpoint from high-resolution complete point clouds.
Afterwards, we downsample incomplete point clouds to the same points (2,048 points) by using farthest distance sampling.
Two missing ratios, 25\% and 50\%, are used in MVP-40 dataset.
Notably, partial and complete point clouds can have different points, since they are sampled by two different processes.

\noindent
\textbf{Completion Evaluation.}
The completion performance for partial point clouds with 25\% and 50\% missing ratios by various methods are reported in Table~\ref{tab:mvp40_cp}.
{Note that we train different models for the data with different missing ratios.}
For both missing ratios, \OM{} shows much better completion results than the other methods.
Comparing completion results by different missing ratios, a larger missing ratio makes the point cloud completion more challenging, but \OM{} even achieves better completion for 50\% missing ratio (CD $=2.76\times10^{-3}$, F1 $=0.776$) than those by PCN dealing with partial point clouds with 25\% missing ratio (CD $=3.87\times10^{-3}$, F1 $=0.577$).

{
\setlength{\tabcolsep}{4.75pt}
\begin{table}[t]
    \caption{Completion results (CD $\times10^3$) on MVP-40 (2,048 points).}
    \vspace{-4mm}
    \begin{center}
    \small{
        \begin{tabular}{l|cc|cc|cc|cc}
            \Xhline{1pt}
            \multirow{2}{*}{{\small{Method}}} & \multicolumn{2}{c|}{\scriptsize{PCN~\cite{yuan2018pcn}}} & \multicolumn{2}{c|}{\scriptsize{TopNet~\cite{tchapmi2019topnet}}} & \multicolumn{2}{c|}{\scriptsize{ECG~\cite{pan2020ecg}}} & \multicolumn{2}{c}{\scriptsize{VRCNet (Ours)}} \\
            \cline{2-9}
             & \scriptsize{CD} & \scriptsize{F1} & \scriptsize{CD} & \scriptsize{F1} & \scriptsize{CD} & \scriptsize{F1} & \scriptsize{CD} & \scriptsize{F1} 
            \\
             \hline\hline
            \footnotesize{25\%} & \scriptsize{3.87} & \scriptsize{0.577} & \scriptsize{3.76} & \scriptsize{0.519} & \scriptsize{2.19} & \scriptsize{0.741} & \scriptsize{\textbf{1.89}} & \scriptsize{\textbf{0.785}} \\
            \footnotesize{50\%} & \scriptsize{4.41} & \scriptsize{0.558} & \scriptsize{5.22} & \scriptsize{0.485} & \scriptsize{3.00} & \scriptsize{0.713} & \scriptsize{\textbf{2.76}} & \scriptsize{\textbf{0.776}} \\
            \Xhline{1pt}
            
        \end{tabular}
    }    
    \end{center}
    \vspace{-4mm}
    \label{tab:mvp40_cp}
\end{table}
}
{
\setlength{\tabcolsep}{4.4pt}
\begin{table}[t]
    \caption{Classification results on MVP-40 dataset (2,048 points).}
    \vspace{-4mm}
    \begin{center}
    \small{
        \begin{tabular}{l|cc|cc|cc|cc}
            \Xhline{1pt}
            \multirow{3}{*}{{\small{Method}}} & \multicolumn{4}{c|}{{\footnotesize{PointNet~\cite{qi2017pointnet}}}} & \multicolumn{4}{c}{{\footnotesize{DGCNN~\cite{dgcnn}}}} \\
            \cline{2-9}
             & \multicolumn{2}{c|}{\scriptsize{Acc. (\%)}} & \multicolumn{2}{c|}{\scriptsize{Avg. (\%)}} & \multicolumn{2}{c|}{\scriptsize{Acc. (\%)}} & \multicolumn{2}{c}{\scriptsize{Avg. (\%)}} \\
            \cline{2-9}
             & \scriptsize{25\%} & \scriptsize{50\%} & \scriptsize{25\%} & \scriptsize{50\%} & \scriptsize{25\%} & \scriptsize{50\%} & \scriptsize{25\%} & \scriptsize{50\%} \\
             \hline\hline
            \scriptsize{Partial} & \scriptsize{58.9} & \scriptsize{28.5} & \scriptsize{58.0} & \scriptsize{29.0} & {\scriptsize{69.0}} & \scriptsize{35.6} & {\scriptsize{66.8}} & {\scriptsize{34.6}} \\
            \; \scriptsize{+ PCN~\cite{yuan2018pcn}} & {\scriptsize{74.9}} & \scriptsize{68.1} & {\scriptsize{71.6}} & \scriptsize{64.2} & \scriptsize{64.9} & \scriptsize{65.7} & \scriptsize{59.9} & \scriptsize{60.1} \\
            \; \scriptsize{+ TopNet~\cite{tchapmi2019topnet}} & \scriptsize{71.3} & \scriptsize{60.5} & \scriptsize{69.3} & \scriptsize{57.6} & \scriptsize{49.3} & \scriptsize{48.6} & \scriptsize{45.9} & \scriptsize{43.7} \\
            \; \scriptsize{+ ECG~\cite{pan2020ecg}} & \scriptsize{74.4} & \textbf{\scriptsize{68.9}} & \scriptsize{70.8} & \textbf{\scriptsize{65.0}} & \scriptsize{70.7} & \scriptsize{68.0} & \scriptsize{65.6} & \scriptsize{62.5} \\
            \; \scriptsize{+ VRCNet (Ours)} & \textbf{\scriptsize{75.7}} & {\scriptsize{67.7}} & \textbf{\scriptsize{73.1}} & {\scriptsize{63.9}} & \textbf{\scriptsize{75.9}} & \textbf{\scriptsize{72.4}} & \textbf{\scriptsize{71.3}} & \textbf{\scriptsize{66.4}} \\
            \hline
            \scriptsize{Complete} & \multicolumn{2}{c|}{\scriptsize{81.1}} & \multicolumn{2}{c|}{\scriptsize{78.6}} & \multicolumn{2}{c|}{\scriptsize{84.9}} & \multicolumn{2}{c}{\scriptsize{83.1}} \\
            \Xhline{1pt}
            
        \end{tabular}
    }    
    \end{center}
    \vspace{-4mm}
    \label{tab:mvp40_cf}
\end{table}
}

\noindent
\textbf{Classification Evaluation.}
Similar with classification on MVP, we train the two methods, PointNet and DGCNN, on complete point clouds from the MVP-40 training set.
Thereafter, we evaluate the trained models on partial point clouds from the MVP-40 test set with two different missing ratios, 25\% and 50\%.
We report classification performance on MVP-40 dataset in Table~\ref{tab:mvp40_cf}.
The classification accuracy decreases as the missing ratio increases, and a relatively small missing ratio (25\%) can cause an obvious classification accuracy drop.
Classification accuracy can be increased after completion, which however does not hold for all cases.
For example, DGCNN achieves Avg. $=69.0$ when classifying partial point clouds with 25\% missing, and using completion methods, such as PCN and TopNet, do not increase its Avg. accuracy.
The reasons can be twofold: on the one side, DGCNN is very sensitive to local point distributions; on the other side, those completion methods cannot recover fine-grained geometric details.
Furthermore, a large missing ratio (50\%) gives rise to a large accuracy drop for both PointNet and DGCNN. 
After completion by \OM{}, the classification accuracy can be highly improved for most cases, especially for 50\% missing ratio.

For complete point cloud classification, PointNet (Acc. $=81.1$, Avg. $=78.6$) performs worse than DGCNN (Acc. $=84.9$, Avg. $=83.1$), but PointNet can outperform DGCNN on those completion results that cannot reconstruct local geometric structures.
For example, PointNet achieves Acc. $=74.9$ and Avg. $=71.6$, while DGCNN only achieves Acc. $=64.9$ and Avg. $=59.9$ on classifying PCN completion results (25\%).
The main reason is that PCN focuses on generating global skeletons but overlooks local details, and DGCNN is more sensitive on local geometric details.
In contrast, DGCNN achieves favorable or on par performance with PointNet on the completion results by \OM{}, which reveals that \OM{} considers both global skeletons and local point distributions for completion.
Furthermore, the per-category classification results (Acc \%) for \OM{} completion by PointNet ({\color[RGB]{238,71,110}\textbf{red}} bar) and DGCNN ({\color[RGB]{15,138,177}\textbf{blue}} bar) are shown in Fig.~\ref{fig:all_cat}.
After completion by \OM{}, both PointNet and DGCNN significantly improve their performance ({\color[RGB]{255,208,102}\textbf{yellow}} bar) for most categories, though slight performance drops ({\color[RGB]{112, 173, 71}\textbf{green}} bar) are observed in few ambiguous categories, such as stairs and TV stand.

\subsection{Completion on Real-world Partial Scans}
\begin{figure}[t!]
    \centering
    \includegraphics[width=1\linewidth]{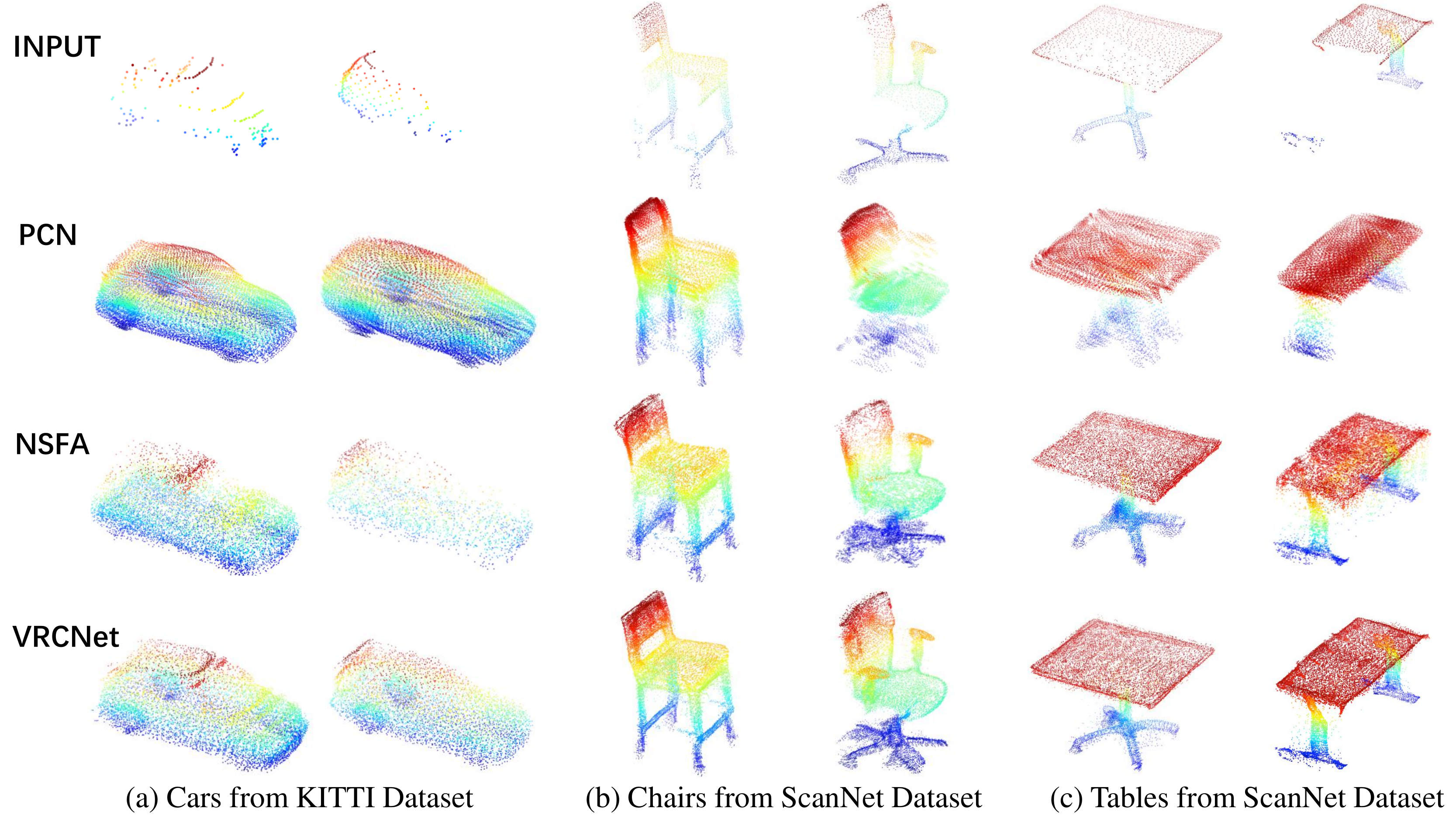}
    \vspace{-6mm}
    \caption{
    \textbf{Completion results on real-scanned point clouds.}   
    \OM{} generates impressive complete shapes for real-scanned point clouds by learning and predicting shape symmetries.
    (a) shows completion results for cars from Kitti dataset~\cite{Geiger2012CVPR}. 
    (b) and (c) show completion results for chairs and tables from ScanNet dataset~\cite{dai2017scannet}, respectively.}
    \label{fig:real_scan}
    \vspace{-3mm}
\end{figure}

We further evaluate \OM{} (trained on MVP with all categories) on real scans, including cars from the KITTI~\cite{Geiger2012CVPR} dataset, chairs and tables from the ScanNet dataset~\cite{dai2017scannet}.
It is noteworthy that the KITTI dataset captured point clouds by using a LiDAR whereas the ScanNet dataset uses a depth camera.
For sparse LiDAR data, we fine-tune all trained models on ShapeNet-car dataset, but no fine-tuning is needed for chairs and tables.
The qualitative completion results are shown in Fig.~\ref{fig:real_scan}.
For those sparse point clouds of cars, \OM{} can predict complete and smooth surfaces that also preserves the observed shape details.  
In comparison, PCN~\cite{yuan2018pcn} suffers a loss of fine shape details and NSFA~\cite{zhang2020detail} cannot generate high-quality complete shapes due to large missing ratios.
For those incomplete chairs and tables, \OM{} generates appealing complete point clouds by exploiting the shape symmetries in the partial scans.

{
\setlength{\tabcolsep}{13pt}
\begin{table}[t]
\caption{A user study of completion quality on real scans. The values are average scores given by volunteers (3 points for best result, 1 point for the worst result). \OM{} is the most preferred method overall.}
\vspace{-4mm}
\begin{center}
\begin{tabular}{l|c|c|c}
\Xhline{1pt}
\small Category & \scriptsize PCN~\cite{yuan2018pcn} & \scriptsize NSFA~\cite{zhang2020detail} & \scriptsize \OM{} \\
\hline\hline
\scriptsize Car (KITTI)      & \scriptsize \textbf{2.87} & \scriptsize 1.07 & \scriptsize 2.07  \\
\scriptsize Chair (ScanNet)   & \scriptsize 1.60  & \scriptsize 1.73 & \scriptsize \textbf{2.67}  \\
\scriptsize Table (ScanNet)    & \scriptsize 1.27  & \scriptsize 2.20 & \scriptsize \textbf{2.60}   \\
\hline
\footnotesize Overall & \scriptsize 1.91 & \scriptsize 1.67 & \scriptsize \textbf{2.45} \\
\Xhline{1pt}
\end{tabular}
\end{center}
\label{tab:user_study}
\end{table}
}

\noindent\textbf{User Study.}
We conduct a user study on the performances of various methods in Tab \ref{tab:user_study}. 
Specifically, we gather a group of 15 volunteers to rank the quality  of complete point cloud predicted by PCN, NSFA, and our \OM, on the real scans of three object categories: car, chair and table. 
For each object category, the volunteers are given three anonymous groups of results, produced by three methods. 
The volunteers are instructed to give the best, middle, and worst results 3, 2, and 1 point(s) respectively. 
We then compute the average scores of all volunteers for each method and class category. 
The evaluation is conducted in a double-blind manner (the methods are anonymous to both the instructor and the volunteers) and the order of the groups are shuffled for each category. 
Our \OM{} is the most favored method overall amongst the three. 
PCN obtains higher score for car completion because it generates smooth mean shapes for all cars, even though few observed shape details of those cars are preserved in their completion results.
However, lacking local details by PCN for scanned cars are easily overlooked by the volunteers, which makes PCN receive the highest score for reconstructing complete cars. 
For the other two categories, chair and table, \OM{} receives the highest scores due to its effectiveness on reconstructing complete shapes by recovering local geometric details using predicted relational structures (\eg shape symmetries).
Nonetheless, overlooking local structures makes PCN achieve the lowest scores on completing chairs and tables.

\section{Conclusion}
In this paper, we propose \OM{}, a variational relational point completion network, which effectively exploits 3D structural relations to predict complete shapes.
Novel self-attention modules, such as 
PSA and PSK,
are proposed for adaptively learning point cloud features, which can be conveniently used in other point cloud tasks.
In addition, we contribute large-scale multi-view partial point cloud datasets, MVP and MVP-40, which totally consist of over 200,000 high-quality 3D point clouds.
Moreover, we perform classification on completion results, which not only validates the qualities of predicted complete point clouds, but also shed some light on robust 3D perception for incomplete real-scans.
We highly encourage researchers to use our proposed novel modules and the MVP dataset for future studies on partial point clouds.




\ifCLASSOPTIONcompsoc
\section*{Acknowledgements}
\else
\section*{Acknowledgement}
\fi

We want to thank Mr. Ziyuan Huang for the valuable discussions.
This study is supported by the Ministry of Education, Singapore, under its MOE AcRF Tier 2 (MOE-T2EP20221- 0012), NTU NAP, and under the RIE2020 Industry Alignment Fund – Industry Collaboration Projects (IAF-ICP) Funding Initiative, as well as cash and in-kind contribution from the industry partner(s).
\ifCLASSOPTIONcaptionsoff
  \newpage
\fi

{
\bibliographystyle{IEEEtran}
\bibliography{egbib}
}


\end{document}